\documentclass{article}
\usepackage[final]{neurips_2024}

\usepackage[utf8]{inputenc} 
\usepackage[T1]{fontenc}    
\usepackage{hyperref}       
\usepackage{url}            
\usepackage{booktabs}       
\usepackage{amsfonts}       
\usepackage{nicefrac}       
\usepackage{microtype}      
\usepackage{xcolor}         

\usepackage{comment}
\usepackage{lipsum}
\usepackage{times}
\usepackage{latexsym}
\usepackage[T1]{fontenc}
\usepackage[utf8]{inputenc}
\usepackage{microtype}
\usepackage{inconsolata}
\usepackage{lipsum}
\usepackage{stfloats}
\usepackage{graphicx}
\usepackage{paralist}
\usepackage{amsmath}
\usepackage{multirow}
\usepackage{caption}
\usepackage{subcaption}
\usepackage{xspace}
\usepackage{makecell}
\usepackage{algorithm}  
\usepackage{algorithmic}
\usepackage{listings}
\usepackage{url}
\usepackage{booktabs}
\usepackage{colortbl}
\usepackage{bbm}
\usepackage{amssymb}
\usepackage{float}
\usepackage[capitalise,noabbrev]{cleveref}

\newcommand{\nonumfootnotetext}[1]{%
  \begingroup
    \renewcommand{\thefootnote}{}%
    \footnotetext{#1}%
  \endgroup
}

\title{Analysing the Residual Stream of Language Models Under Knowledge Conflicts}

\usepackage{marvosym}
\author{%
Yu Zhao$^1$\quad Xiaotang Du$^1$\quad Giwon Hong$^1$\quad  Aryo Pradipta Gema$^1$\quad Alessio Devoto$^3$ \\ \bf \quad Hongru Wang$^2$\quad Xuanli He$^4$\quad Kam-Fai Wong$^2$\quad Pasquale Minervini$^{1,5}$ \\
$^1$University of Edinburgh \quad
$^2$The Chinese University of Hong Kong \\
$^3$Sapienza University of Rome \quad
$^4$University College London \quad $^5$Miniml.AI\\
\texttt{\{yu.zhao, p.minervini\}@ed.ac.uk}\\
}
\begin{document}
\maketitle

\begin{abstract}
Large language models (LLMs) can store a significant amount of factual knowledge in their parameters.
However, their parametric knowledge may conflict with the information provided in the context.
Such conflicts can lead to undesirable model behaviour, such as reliance on outdated or incorrect information.
In this work, we investigate whether LLMs can identify knowledge conflicts and whether it is possible to know which source of knowledge the model will rely on by analysing the residual stream of the LLM.
Through probing tasks, we find that LLMs can internally register the signal of knowledge conflict in the residual stream, which can be accurately detected by probing the intermediate model activations.
This allows us to detect conflicts within the residual stream before generating the answers without modifying the input or model parameters.
Moreover, we find that the residual stream shows significantly different patterns when the model relies on contextual knowledge versus parametric knowledge to resolve conflicts.
This pattern can be employed to estimate the behaviour of LLMs when conflict happens and prevent unexpected answers before producing the answers.
Our analysis offers insights into how LLMs internally manage knowledge conflicts and provides a foundation for developing methods to control the knowledge selection processes.
\nonumfootnotetext{Accepted at the Foundation Model Interventions Workshop @ NeurIPS 2024. This work is the preliminary study of "Steering Knowledge Selection Behaviours in LLMs via SAE-Based Representation Engineering"}
\end{abstract}

\section{Introduction}

Large language models (LLMs) have shown remarkable capability to memorise factual knowledge and solve knowledge-intensive tasks~\citep{DBLP:conf/emnlp/PetroniRRLBWM19,gpt3,llama2,mistral,gemini}.
Nevertheless, the knowledge stored in their parameters (\emph{parametric knowledge}) can be inaccurate or outdated.
To alleviate this issue, retrieval and tool-augmented approaches have been widely adopted to provide LLMs with external knowledge (\emph{contextual knowledge})~\citep{DBLP:conf/emnlp/KarpukhinOMLWEC20,lewis2020retrieval,DBLP:conf/emnlp/WuZHMS022,schick2024toolformer}. 
However, such contextual knowledge can include information that conflicts with the parametric knowledge of the model, which may result in undesired behaviour; for example, the model can rely on inaccurate information sources and produce inaccurate generations~\citep{when-not-to-trust-language-models,conflictqa,conflictbank,resolving-knowledge-conflict,macnoise, zhao2024attacks}.
Prior research found that LLMs tend to prefer contextual knowledge (e.g. retrieved passages) over their parametric knowledge~\citep{conflictbank,conflictqa}.
However, in more general applications, LLMs should retain the ability to use parametric knowledge when presented with incorrect or undesirable information~\citep{chen2023combating,chen2023can,zou2024poisonedrag,when-not-to-trust-language-models,zhong2023poisoning}.
To achieve this goal, LLMs are expected to acknowledge the existence of conflicts, allowing them to alert the user while keeping the decision-making process under the user's control for further action.
Existing works investigate the fine-tuning and prompting-based strategies to detect knowledge conflicts~\citep{resolving-knowledge-conflict}.
These methods need additional interactions with the model, e.g., by asking the LLMs to examine the conflicts sentence by sentence~\citep{resolving-knowledge-conflict}, which may result in high latency times and prevent practical applications of these models.
Additionally, they do not provide insight into how LLMs internally detect and resolve conflicts.
In this work, we analyse the residual stream~\citep{elhage2021mathematical,induction-head} in LLMs to better understand their behaviour when knowledge conflicts arise, especially between parametric knowledge and contextual knowledge.
Our probing experiments on the residual stream indicate that the signal of knowledge conflict rises from the intermediate layers (e.g., the 13th layer of Llama3-8B). 
Utilising this signal, a simple logistic regression model can achieve 90\% accuracy in knowledge conflict detection without modifying the input and parameters of LLMs while introducing only a negligible computation overhead.
Moreover, we also observe that the residual stream exhibits different patterns starting from the middle layers (e.g., the 17th layers of Llama3-8B) when the model takes different source information to resolve the conflict.
For example, when the model uses contextual knowledge, the residual stream exhibits a significantly more skewed distribution compared with when it uses its parametric knowledge.
In conclusion, our analysis of the residual stream reveals that: 1) LLMs exhibit internal mechanisms for identifying conflicts, and this signal can be leveraged to detect conflicts effectively in the mid-layers of LLMs; 2) LLMs display distinct skewness patterns in the residual stream when using different sources of information, which provides insights on the model's behaviour.

\section{Background and Methods}

\paragraph{Residual Stream}
We examine the Transformer architecture from the perspective of the residual stream~\citep{elhage2021mathematical,induction-head}. In this framework, tokens flow through the model, with their embeddings being modified by vector additions from the attention and feed-forward blocks in each layer.
We denote the hidden states at position $i$ at $l$-th layer as $\mathbf{h}_i^l\in \mathbb{R}^{d}$, where $d$ is the dimension of the internal states of the model.
The model produces the initial residual stream $\mathbf{h}_i^0$ by applying an embedding matrix to the tokens.
Then, the model modifies the residual stream by a sequence of $L$ layers Transformers, where each Transformer layer consists of a Self-Attention block and MLP at $l$-th layer.
Formally, denote $\mathbf{a}_i^l$ and $\mathbf{m}_i^l$ as the activations of Self-Attention and MLP respectively, the update of the residual stream at $l$-th layer is $\mathbf{h}^{l'} = \text{LayerNorm}(\mathbf{h}^{l-1}) + \mathbf{a}_{i}^{l}$ and $\mathbf{h}^{l}  = \text{LayerNorm}(\mathbf{h}^{l'}) + \mathbf{m}_{i}^{l}$.

\paragraph{Linear Probing} 
Linear probing~\citep{conneau2018you,zhu2023physics,allen2023physics} is a commonly used technique to analyse whether certain information is encoded within the residual stream of a language model.
Specifically, for an activation $\mathbf{x}$ from the residual stream, i.e., $\mathbf{h}$, $\mathbf{a}$, or $\mathbf{m}$, a logistic regression model is applied to perform binary classification: $P(y=1|\mathbf{x}) = \delta\left( \mathbf{x} \mathbf{W} \right)$,
where $\mathbf{W} \in \mathbb{R}^{d \times 1}$ is the learned weight
that linearly projects the activation into a scalar value
, and $\delta$ is the Sigmoid function that outputs the likelihood of probed information existing in the activation.

\section{Experimental Setup}

\label{sec:problem-setip}

\paragraph{Problem Setup} Following previous studies~\citep{nqswap,macnoise,conflictqa,conflictbank,resolving-knowledge-conflict}, we use open-domain question-answering (ODQA) tasks to investigate the behaviours of LLMs when there is a conflict between the model's parametric knowledge and contextual knowledge.
In ODQA datasets with knowledge conflicts, each instance is presented as $(q, e_{M}, e_{C}, a_{M}, a_{C})$, where $q$ is the question, $e_{M}$ is the evidence that supports the memorised knowledge, $e_{C}$ is the evidence that conflicts with the language model's memorised knowledge, $a_{M}$ is the answer based on and $e_{M}$, and $a_{C}$ is the answer based on the $e_{C}$.
The model's parametric knowledge is tested in the close-book setting, where the model generates answer $a_{M}$ based on the question $q$ without external evidence.
We generate the answers using a greedy decoding strategy. We use three in-context demonstrations to align the answer format and, for fairness, use the same in-context demonstrations in all experiments.
\paragraph{Datasets and Models}
We use NQSwap~\citep{nqswap}, Macnoise~\citep{macnoise} and ConflictQA~\citep{conflictqa} to analyse the residual stream when knowledge conflicts arise.
We present the experiment results of NQSwap using Llama3-8B~\citep{llama3} in the main paper, and the results of other datasets and models are provided in \cref{sec:more-conflict-probing} and \cref{sec:more-skewness-plots}. The training details of the probing model are presented in~\cref{sec:probing-training}

\begin{figure}[t]
    \centering
    \begin{subfigure}[b]{0.32\textwidth}
        \centering
        \includegraphics[width=\linewidth]{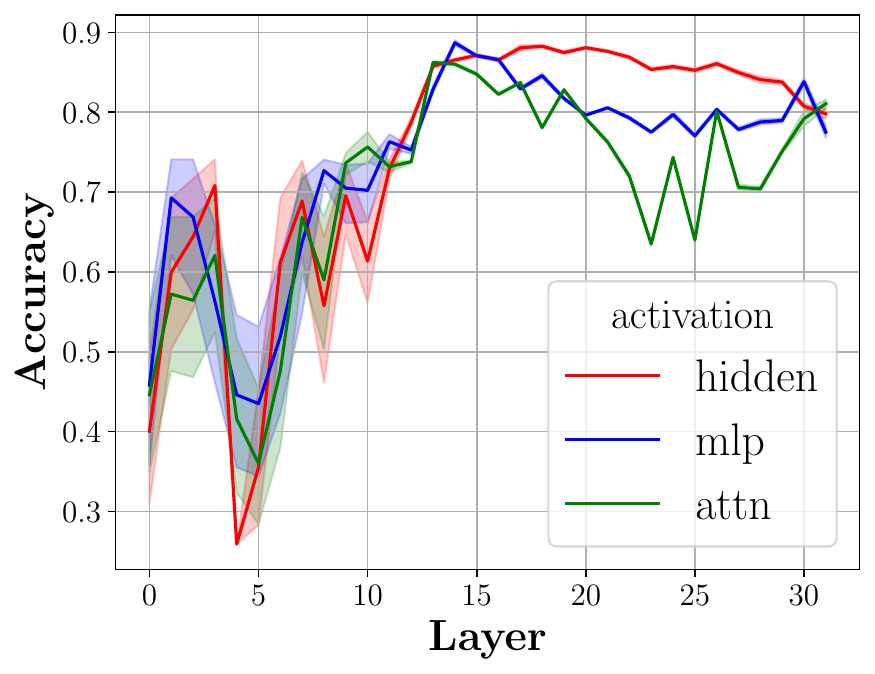}
        \caption{Accuracy}
    \end{subfigure}
    \begin{subfigure}[b]{0.32\textwidth}
        \centering
        \includegraphics[width=\linewidth]{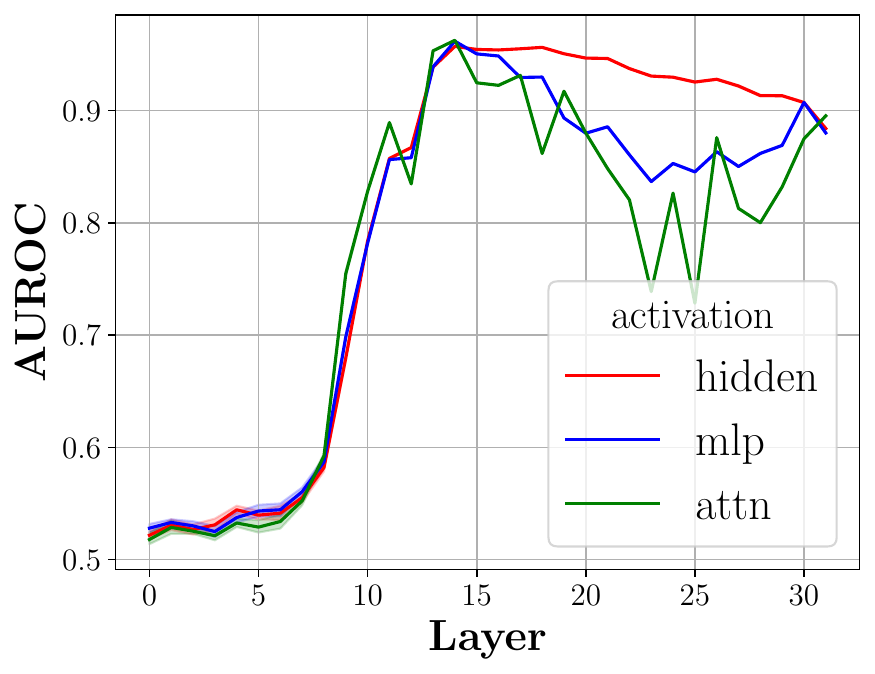}
        \caption{AUROC}
    \end{subfigure}
    \begin{subfigure}[b]{0.32\textwidth}
        \centering
        \includegraphics[width=\linewidth]{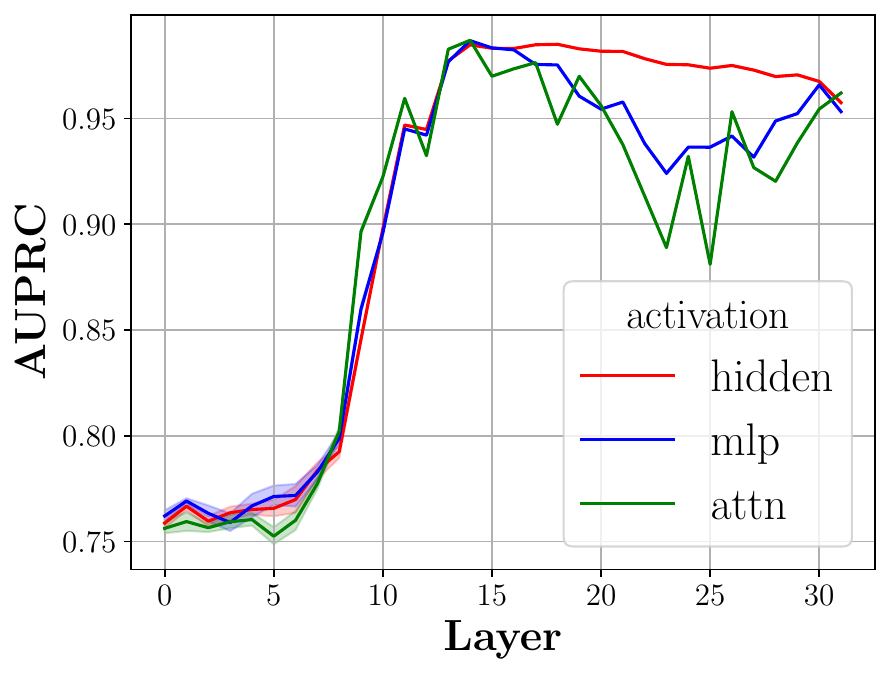}
        \caption{AUPRC}
    \end{subfigure}
\caption{Accuracy, AUROC, and AUPRC of probing models on detecting the knowledge conflicts based on the activations of Llama3-8B. The probing results on hidden state, MLP and Self-Attention activation are coloured red, blue and green, respectively. More analysis is presented in~\cref{sec:more-conflict-probing}.
}
\label{fig:knowledge-conflict-probing-llama3}
\end{figure}

\section{Results and Findings}
\label{sec:results-and-findings}
In this work, we aim to answer the two following research questions:
\textit{1)} Can we identify the conflict between context and parameter knowledge by probing the residual stream?
\textit{2)} Can we know which source of knowledge the models will use before they generate the answers?
We probe and analyse the residual stream to answer these two questions in the following parts.
\paragraph{Identifying Knowledge Conflicts by Probing the Residual Stream}
We analyse whether language models can identify contextual-parametric knowledge conflicts by probing the residual stream.
To this end, we create two groups of instances, $D^{e_C} = \{(q, e_C)\}$ and $D^{e_M} = \{(q, e_M)\}$, where the model generates answers based on conflict evidence in $D^{e_C}$ and non-conflict evidence in $D^{e_M}$.
The probing model is trained to classify whether a given activation is from $D^{e_C}$ or $D^{e_M}$.
We probe the residual stream at the final position to determine if the model is aware of the conflict during the first token generation. This is because the hidden state at the last position in the output layer is used to predict the first token of the answer.
For each activation $\mathbf{h}^l$, $\mathbf{a}^l$ and $\mathbf{m}^l$ at each layer, we train a probing model to classify whether it belongs to $D^{e_M}$ or $D^{e_C}$.

As shown in \cref{fig:knowledge-conflict-probing-llama3}(b) and \cref{fig:knowledge-conflict-probing-llama3}(c), the AUROC and AUPRC of the probing models increase from the first layer to the 14th layer, and this trend is same across the hidden state, MLP, and Self-Attention activations.
In \cref{fig:knowledge-conflict-probing-llama3}(a), the accuracy of the probing models at the early layers is random; similar to the trend of AUROC and AUPRC, the accuracy also reaches the highest score at the 14th layer.
The above observation indicates that the residual stream does not contain information about knowledge conflict at the early layers.
This information rises from around the 8th layer and reaches the highest point at the 14th layer.
After the 14th layer, the probing model's performance decreases slightly until the last layer.
Besides, we also observe that the probing results of MLP and Self-Attention activations show a significantly lower accuracy than the hidden state after the 14th layer, which may suggest that MLP and Self-Attention do not provide further conflicting information into the residual stream.
We find the same trend using Llama2-7B as shown in~\cref{fig:knowledge-conflict-probing-llama2}.

\paragraph{Analysis of the Residual Stream When LLMs Using Different Sources of Knowledge}
We investigate the distribution patterns of the residual stream when the language model uses different sources of information to generate the answer.
Based on the model's predictions on instances belongs to $D^{e_C}$, we classify them into two groups: $D_{a_C}^{e_C}$ and $D_{a_M}^{e_C}$.
Here, $D_{a_C}^{e_C}$ represents the set of instances where the model's predictions align with $a_C$, while $D_{a_M}^{e_C}$ contains the instances where the predictions align with $a_M$.
The model uses contextual knowledge and parametric knowledge to answer the questions from $D_{a_C}^{e_C}$ and $D_{a_M}^{e_C}$, respectively.

\begin{figure}[t]
    \centering
    \begin{subfigure}[b]{0.32\textwidth}
        \centering
        \includegraphics[width=\linewidth]{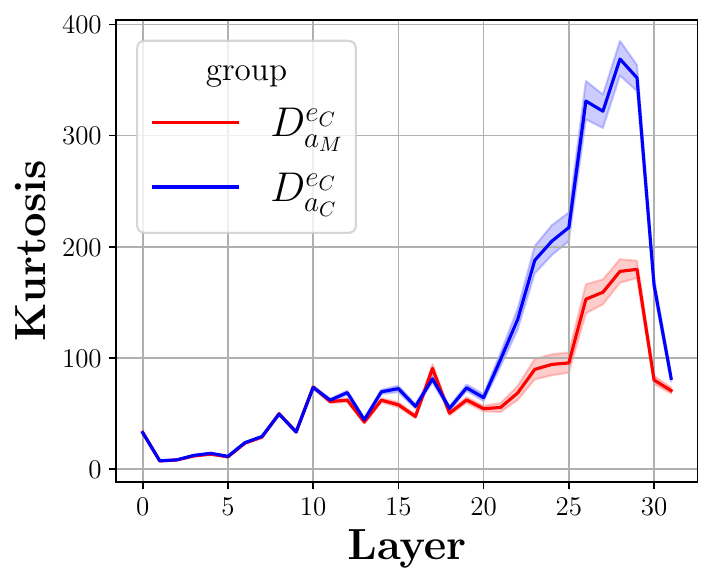}
    \end{subfigure}
    \begin{subfigure}[b]{0.32\textwidth}
        \centering
        \includegraphics[width=\linewidth]{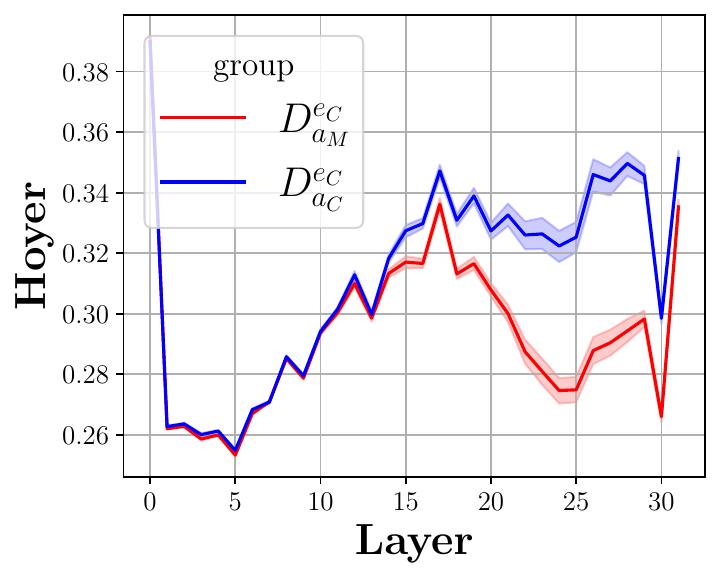}
    \end{subfigure}
    \begin{subfigure}[b]{0.32\textwidth}
        \centering
        \includegraphics[width=\linewidth]{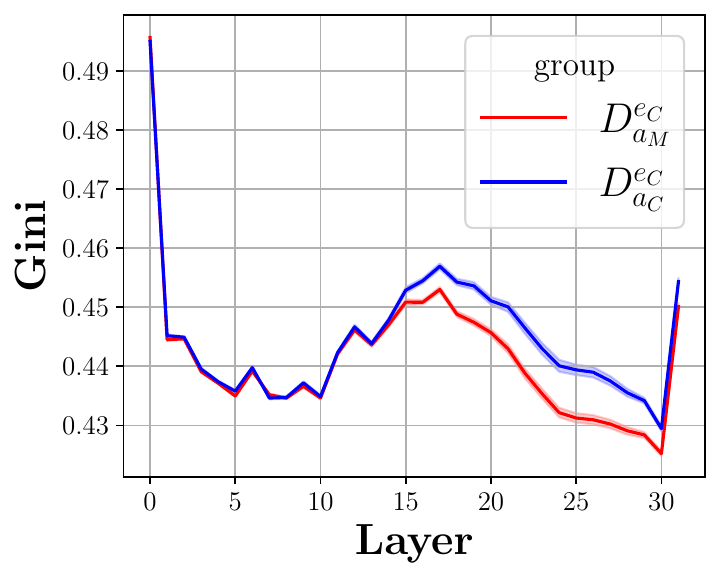}
    \end{subfigure}
\caption{Skewness of the hidden state activations of Llama3-8B when in presence of knowledge conflicts. 
Blue and red lines represent the skewness of hidden states from $D_{a_C}^{e_C}$ and $D_{a_M}^{e_C}$, respectively.
Higher scores indicate a more skewed distribution. Additional analyses are available in~\cref{sec:more-skewness-plots}.
}
\label{fig:kurtosis-llama3}
\end{figure}

\begin{figure}[t]
    \centering
    \begin{subfigure}[b]{0.32\textwidth}
        \centering
        \includegraphics[width=\linewidth]{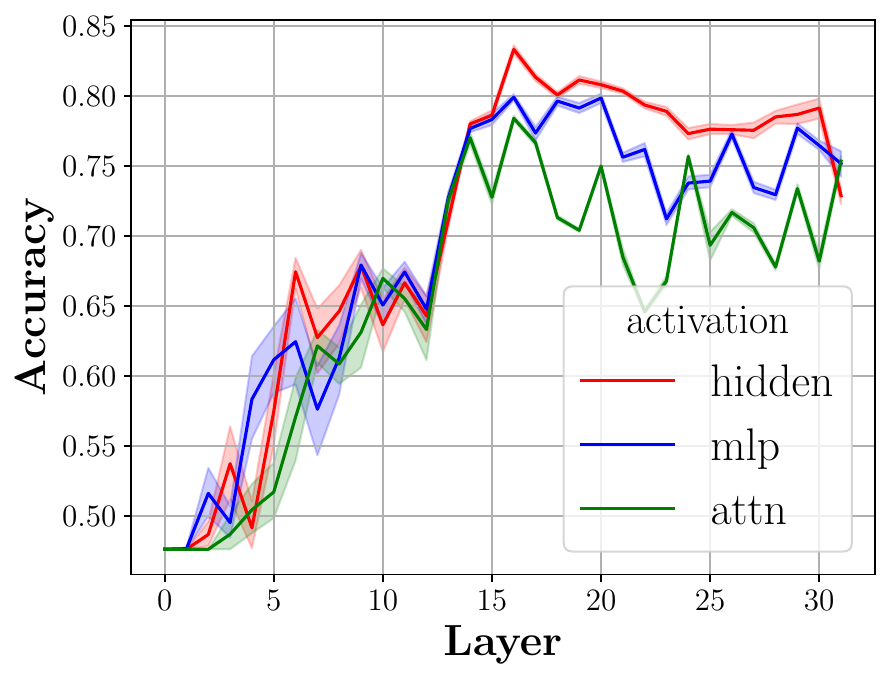}
    \end{subfigure}
    \begin{subfigure}[b]{0.32\textwidth}
        \centering
        \includegraphics[width=\linewidth]{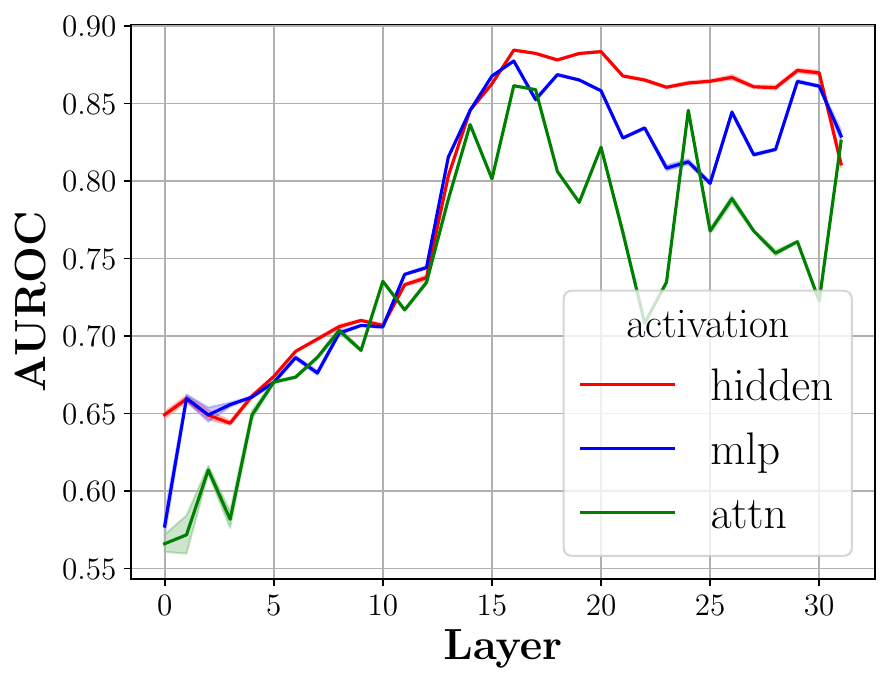}
    \end{subfigure}
    \begin{subfigure}[b]{0.32\textwidth}
        \centering
        \includegraphics[width=\linewidth]{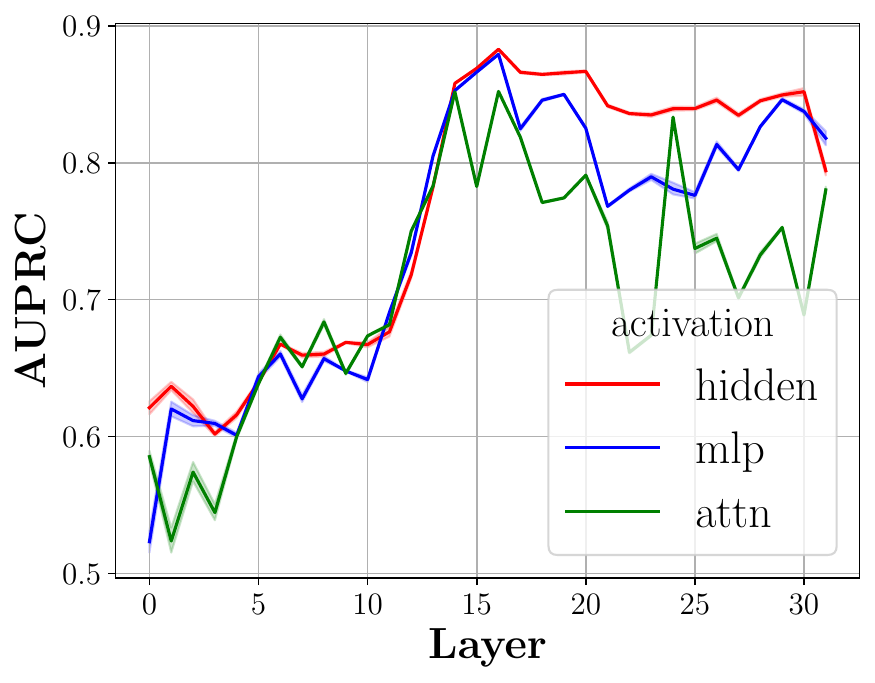}
    \end{subfigure}
\caption{Accuracy, AUROC, and AUPRC of probing models on predicting which source of knowledge the model will use to predict the answer in Llama3-8B. More results are Skewness of the hidden state activations of Llama3-8B when the model uses knowledge from different sources to predict the answer.  Additional results are available in~\cref{sec:more-behaviour}.
}
\label{fig:behaviour-probing-llama3}
\end{figure}

First, we examine the residual streams' distribution patterns in the two groups of instances $D_{a_C}^{e_C}$ and $D_{a_M}^{e_C}$. 
We measure the skewness of the residual stream using Kurtosis, Hoyer and Gini index.
We present the results of NQSwap using Llama3-8B in~\cref{fig:kurtosis-llama3}, and more results are provided in the \cref{sec:more-skewness-plots}.
We find that when the model uses contextual knowledge for prediction ($D_{a_C}^{e_C}$, blue lines shown in \cref{fig:kurtosis-llama3}), the residual stream shows a significantly skewed distribution compared with using parametric knowledge from the 20th to 30th layers.
Therefore, the distribution patterns of the residual stream can indicate the model will use different sources of knowledge.
It provides the foundation for predicting the model's behaviour in advance, which can be used to mitigate the generation of undesirable responses in advance.

Based on the above observation, we probe the residual stream to analyse the possibility of predicting which source of knowledge will be used to generate the answer.
The probing model is trained to classify whether the model will generate $a_C$ or $a_M$ based on the activation from $D_{a_C}^{e_C}$ or $D_{a_M}^{e_C}$.
We present the probing results in ~\cref{fig:behaviour-probing-llama3}.
We observe that the probing model's performance gradually improves from the first layer to the 16th layer, which occurs after the signal of knowledge conflict has already reached its peak at the 13th and 14th layers.
This observation suggests that the decision of which knowledge to use occurs after the detection of the knowledge conflict signal.

\section{Related Work}

Contextual and parametric knowledge conflict can happen when the retrieved external knowledge in the context does not agree with the parametric knowledge which is memorised during pre-training~\citep{nqswap, xu2024knowledge, conflictqa, conflictbank, resolving-knowledge-conflict, when-not-to-trust-language-models}.
Previous works found models may prefer the contextual knowledge~\citep{resolving-knowledge-conflict, conflictbank, conflictqa, competition-of-mechanisms} when parametric and contextual knowledge conflicts, and the relevance, length, and the number of the evidence will influence the model's preferences~\citep{conflictqa, conflictbank}.
To detect the conflict, previous work~\citep{resolving-knowledge-conflict} designed a multi-step prompting strategy to detect the knowledge, which involves parametric knowledge generation, fine-grained sentence consistency checking, and potential conflict reduction.
However, this pipeline significantly reduces efficiency and lacks an understanding of the mechanism of how LLMs detect and resolve conflict.

\section{Conclusions}
In this work, we analyse the residual stream of the language models when context-parameter knowledge conflicts.
First, we find that LLMs exhibit internal mechanisms for identifying conflicts in the mid-layers.
Second, we find that the residual stream shows distinct skewness patterns when the model uses context and parametric knowledge to predict.
Our analysis provides insights into the behaviour of LLMs in the presence of knowledge conflicts.
This work is the preliminary study of "Steering Knowledge Selection Behaviours in LLMs via SAE-Based Representation Engineering~\citep{zhao2024steering}", where we propose a training-free inference-time intervention method \textsc{SpARE} to steer the knowledge selection behaviours of LLMs under context-memory knowledge conflict.

\section*{Acknowledgements}
Yu Zhao and Xiaotang Du were partly supported by the UKRI Centre for Doctoral Training in Natural Language Processing, funded by UK Research and Innovation (grant EP/S022481/1) and the University of Edinburgh, School of Informatics.
Giwon Hong was supported by the ILCC PhD program (School of Informatics Funding Package) at the University of Edinburgh, School of Informatics.
Aryo Pradipta Gema was supported by the United Kingdom Research and Innovation (grant EP/S02431X/1), UKRI Centre for Doctoral Training in Biomedical AI at the University of Edinburgh, School of Informatics.
Alessio Devoto was supported by Sapienza Grant RM1221816BD028D6 (DeSMOS).
Xuanli He was funded by an industry grant from Cisco.
Pasquale Minervini was partially funded by ELIAI (The Edinburgh Laboratory for Integrated Artificial Intelligence), EPSRC (grant no.\ EP/W002876/1), an industry grant from Cisco, and a donation from Accenture LLP.

\bibliography{reference}

\begin{thebibliography}{29}
\providecommand{\natexlab}[1]{#1}
\providecommand{\url}[1]{\texttt{#1}}
\expandafter\ifx\csname urlstyle\endcsname\relax
  \providecommand{\doi}[1]{doi: #1}\else
  \providecommand{\doi}{doi: \begingroup \urlstyle{rm}\Url}\fi

\bibitem[Allen-Zhu and Li(2023)]{allen2023physics}
Zeyuan Allen-Zhu and Yuanzhi Li.
\newblock Physics of language models: Part 1, context-free grammar.
\newblock \emph{arXiv preprint arXiv:2305.13673}, 2023.

\bibitem[Brown(2020)]{gpt3}
Tom~B Brown.
\newblock Language models are few-shot learners.
\newblock \emph{arXiv preprint arXiv:2005.14165}, 2020.

\bibitem[Chen and Shu(2023{\natexlab{a}})]{chen2023can}
Canyu Chen and Kai Shu.
\newblock Can llm-generated misinformation be detected?
\newblock \emph{arXiv preprint arXiv:2309.13788}, 2023{\natexlab{a}}.

\bibitem[Chen and Shu(2023{\natexlab{b}})]{chen2023combating}
Canyu Chen and Kai Shu.
\newblock Combating misinformation in the age of llms: Opportunities and challenges.
\newblock \emph{AI Magazine}, 2023{\natexlab{b}}.

\bibitem[Conneau et~al.(2018)Conneau, Kruszewski, Lample, Barrault, and Baroni]{conneau2018you}
Alexis Conneau, German Kruszewski, Guillaume Lample, Lo{\"\i}c Barrault, and Marco Baroni.
\newblock What you can cram into a single vector: Probing sentence embeddings for linguistic properties.
\newblock \emph{arXiv preprint arXiv:1805.01070}, 2018.

\bibitem[Dubey et~al.(2024)Dubey, Jauhri, Pandey, Kadian, Al-Dahle, Letman, Mathur, Schelten, Yang, Fan, et~al.]{llama3}
Abhimanyu Dubey, Abhinav Jauhri, Abhinav Pandey, Abhishek Kadian, Ahmad Al-Dahle, Aiesha Letman, Akhil Mathur, Alan Schelten, Amy Yang, Angela Fan, et~al.
\newblock The llama 3 herd of models.
\newblock \emph{arXiv preprint arXiv:2407.21783}, 2024.

\bibitem[Elhage et~al.(2021)Elhage, Nanda, Olsson, Henighan, Joseph, Mann, Askell, Bai, Chen, Conerly, et~al.]{elhage2021mathematical}
Nelson Elhage, Neel Nanda, Catherine Olsson, Tom Henighan, Nicholas Joseph, Ben Mann, Amanda Askell, Yuntao Bai, Anna Chen, Tom Conerly, et~al.
\newblock A mathematical framework for transformer circuits.
\newblock \emph{Transformer Circuits Thread}, 1\penalty0 (1):\penalty0 12, 2021.

\bibitem[Hong et~al.(2024)Hong, Kim, Kang, Myaeng, and Whang]{macnoise}
Giwon Hong, Jeonghwan Kim, Junmo Kang, Sung{-}Hyon Myaeng, and Joyce~Jiyoung Whang.
\newblock Why so gullible? enhancing the robustness of retrieval-augmented models against counterfactual noise.
\newblock In Kevin Duh, Helena G{\'{o}}mez{-}Adorno, and Steven Bethard, editors, \emph{Findings of the Association for Computational Linguistics: {NAACL} 2024, Mexico City, Mexico, June 16-21, 2024}, pages 2474--2495. Association for Computational Linguistics, 2024.
\newblock \doi{10.18653/V1/2024.FINDINGS-NAACL.159}.
\newblock URL \url{https://doi.org/10.18653/v1/2024.findings-naacl.159}.

\bibitem[Jiang et~al.(2023)Jiang, Sablayrolles, Mensch, Bamford, Chaplot, Casas, Bressand, Lengyel, Lample, Saulnier, et~al.]{mistral}
Albert~Q Jiang, Alexandre Sablayrolles, Arthur Mensch, Chris Bamford, Devendra~Singh Chaplot, Diego de~las Casas, Florian Bressand, Gianna Lengyel, Guillaume Lample, Lucile Saulnier, et~al.
\newblock Mistral 7b.
\newblock \emph{arXiv preprint arXiv:2310.06825}, 2023.

\bibitem[Karpukhin et~al.(2020)Karpukhin, Oguz, Min, Lewis, Wu, Edunov, Chen, and Yih]{DBLP:conf/emnlp/KarpukhinOMLWEC20}
Vladimir Karpukhin, Barlas Oguz, Sewon Min, Patrick S.~H. Lewis, Ledell Wu, Sergey Edunov, Danqi Chen, and Wen{-}tau Yih.
\newblock Dense passage retrieval for open-domain question answering.
\newblock In Bonnie Webber, Trevor Cohn, Yulan He, and Yang Liu, editors, \emph{Proceedings of the 2020 Conference on Empirical Methods in Natural Language Processing, {EMNLP} 2020, Online, November 16-20, 2020}, pages 6769--6781. Association for Computational Linguistics, 2020.
\newblock \doi{10.18653/V1/2020.EMNLP-MAIN.550}.
\newblock URL \url{https://doi.org/10.18653/v1/2020.emnlp-main.550}.

\bibitem[Lewis et~al.(2020)Lewis, Perez, Piktus, Petroni, Karpukhin, Goyal, K{\"u}ttler, Lewis, Yih, Rockt{\"a}schel, et~al.]{lewis2020retrieval}
Patrick Lewis, Ethan Perez, Aleksandra Piktus, Fabio Petroni, Vladimir Karpukhin, Naman Goyal, Heinrich K{\"u}ttler, Mike Lewis, Wen-tau Yih, Tim Rockt{\"a}schel, et~al.
\newblock Retrieval-augmented generation for knowledge-intensive nlp tasks.
\newblock \emph{Advances in Neural Information Processing Systems}, 33:\penalty0 9459--9474, 2020.

\bibitem[Longpre et~al.(2021)Longpre, Perisetla, Chen, Ramesh, DuBois, and Singh]{nqswap}
Shayne Longpre, Kartik Perisetla, Anthony Chen, Nikhil Ramesh, Chris DuBois, and Sameer Singh.
\newblock Entity-based knowledge conflicts in question answering.
\newblock In Marie-Francine Moens, Xuanjing Huang, Lucia Specia, and Scott Wen-tau Yih, editors, \emph{Proceedings of the 2021 Conference on Empirical Methods in Natural Language Processing}, pages 7052--7063, Online and Punta Cana, Dominican Republic, November 2021. Association for Computational Linguistics.
\newblock \doi{10.18653/v1/2021.emnlp-main.565}.
\newblock URL \url{https://aclanthology.org/2021.emnlp-main.565}.

\bibitem[Mallen et~al.(2023)Mallen, Asai, Zhong, Das, Khashabi, and Hajishirzi]{when-not-to-trust-language-models}
Alex Mallen, Akari Asai, Victor Zhong, Rajarshi Das, Daniel Khashabi, and Hannaneh Hajishirzi.
\newblock When not to trust language models: Investigating effectiveness of parametric and non-parametric memories.
\newblock In Anna Rogers, Jordan~L. Boyd{-}Graber, and Naoaki Okazaki, editors, \emph{Proceedings of the 61st Annual Meeting of the Association for Computational Linguistics (Volume 1: Long Papers), {ACL} 2023, Toronto, Canada, July 9-14, 2023}, pages 9802--9822. Association for Computational Linguistics, 2023.
\newblock \doi{10.18653/V1/2023.ACL-LONG.546}.
\newblock URL \url{https://doi.org/10.18653/v1/2023.acl-long.546}.

\bibitem[Olsson et~al.(2022)Olsson, Elhage, Nanda, Joseph, DasSarma, Henighan, Mann, Askell, Bai, Chen, et~al.]{induction-head}
Catherine Olsson, Nelson Elhage, Neel Nanda, Nicholas Joseph, Nova DasSarma, Tom Henighan, Ben Mann, Amanda Askell, Yuntao Bai, Anna Chen, et~al.
\newblock In-context learning and induction heads.
\newblock \emph{arXiv preprint arXiv:2209.11895}, 2022.

\bibitem[Ortu et~al.(2024)Ortu, Jin, Doimo, Sachan, Cazzaniga, and Sch{\"o}lkopf]{competition-of-mechanisms}
Francesco Ortu, Zhijing Jin, Diego Doimo, Mrinmaya Sachan, Alberto Cazzaniga, and Bernhard Sch{\"o}lkopf.
\newblock Competition of mechanisms: Tracing how language models handle facts and counterfactuals.
\newblock \emph{arXiv preprint arXiv:2402.11655}, 2024.

\bibitem[Petroni et~al.(2019)Petroni, Rockt{\"{a}}schel, Riedel, Lewis, Bakhtin, Wu, and Miller]{DBLP:conf/emnlp/PetroniRRLBWM19}
Fabio Petroni, Tim Rockt{\"{a}}schel, Sebastian Riedel, Patrick S.~H. Lewis, Anton Bakhtin, Yuxiang Wu, and Alexander~H. Miller.
\newblock Language models as knowledge bases?
\newblock In Kentaro Inui, Jing Jiang, Vincent Ng, and Xiaojun Wan, editors, \emph{Proceedings of the 2019 Conference on Empirical Methods in Natural Language Processing and the 9th International Joint Conference on Natural Language Processing, {EMNLP-IJCNLP} 2019, Hong Kong, China, November 3-7, 2019}, pages 2463--2473. Association for Computational Linguistics, 2019.
\newblock \doi{10.18653/V1/D19-1250}.
\newblock URL \url{https://doi.org/10.18653/v1/D19-1250}.

\bibitem[Schick et~al.(2024)Schick, Dwivedi-Yu, Dess{\`\i}, Raileanu, Lomeli, Hambro, Zettlemoyer, Cancedda, and Scialom]{schick2024toolformer}
Timo Schick, Jane Dwivedi-Yu, Roberto Dess{\`\i}, Roberta Raileanu, Maria Lomeli, Eric Hambro, Luke Zettlemoyer, Nicola Cancedda, and Thomas Scialom.
\newblock Toolformer: Language models can teach themselves to use tools.
\newblock \emph{Advances in Neural Information Processing Systems}, 36, 2024.

\bibitem[Su et~al.(2024)Su, Zhang, Qu, Zhu, Li, Sun, Li, Zhang, and Cheng]{conflictbank}
Zhaochen Su, Jun Zhang, Xiaoye Qu, Tong Zhu, Yanshu Li, Jiashuo Sun, Juntao Li, Min Zhang, and Yu~Cheng.
\newblock Conflictbank: A benchmark for evaluating the influence of knowledge conflicts in llm.
\newblock \emph{arXiv preprint arXiv:2408.12076}, 2024.

\bibitem[Team et~al.(2023)Team, Anil, Borgeaud, Wu, Alayrac, Yu, Soricut, Schalkwyk, Dai, Hauth, et~al.]{gemini}
Gemini Team, Rohan Anil, Sebastian Borgeaud, Yonghui Wu, Jean-Baptiste Alayrac, Jiahui Yu, Radu Soricut, Johan Schalkwyk, Andrew~M Dai, Anja Hauth, et~al.
\newblock Gemini: a family of highly capable multimodal models.
\newblock \emph{arXiv preprint arXiv:2312.11805}, 2023.

\bibitem[Touvron et~al.(2023)Touvron, Martin, Stone, Albert, Almahairi, Babaei, Bashlykov, Batra, Bhargava, Bhosale, et~al.]{llama2}
Hugo Touvron, Louis Martin, Kevin Stone, Peter Albert, Amjad Almahairi, Yasmine Babaei, Nikolay Bashlykov, Soumya Batra, Prajjwal Bhargava, Shruti Bhosale, et~al.
\newblock Llama 2: Open foundation and fine-tuned chat models.
\newblock \emph{arXiv preprint arXiv:2307.09288}, 2023.

\bibitem[Wang et~al.(2023)Wang, Feng, Wang, Shi, Balachandran, He, and Tsvetkov]{resolving-knowledge-conflict}
Yike Wang, Shangbin Feng, Heng Wang, Weijia Shi, Vidhisha Balachandran, Tianxing He, and Yulia Tsvetkov.
\newblock Resolving knowledge conflicts in large language models.
\newblock \emph{CoRR}, abs/2310.00935, 2023.
\newblock \doi{10.48550/ARXIV.2310.00935}.
\newblock URL \url{https://doi.org/10.48550/arXiv.2310.00935}.

\bibitem[Wu et~al.(2022)Wu, Zhao, Hu, Minervini, Stenetorp, and Riedel]{DBLP:conf/emnlp/WuZHMS022}
Yuxiang Wu, Yu~Zhao, Baotian Hu, Pasquale Minervini, Pontus Stenetorp, and Sebastian Riedel.
\newblock An efficient memory-augmented transformer for knowledge-intensive {NLP} tasks.
\newblock In Yoav Goldberg, Zornitsa Kozareva, and Yue Zhang, editors, \emph{Proceedings of the 2022 Conference on Empirical Methods in Natural Language Processing, {EMNLP} 2022, Abu Dhabi, United Arab Emirates, December 7-11, 2022}, pages 5184--5196. Association for Computational Linguistics, 2022.
\newblock \doi{10.18653/V1/2022.EMNLP-MAIN.346}.
\newblock URL \url{https://doi.org/10.18653/v1/2022.emnlp-main.346}.

\bibitem[Xie et~al.(2024)Xie, Zhang, Chen, Lou, and Su]{conflictqa}
Jian Xie, Kai Zhang, Jiangjie Chen, Renze Lou, and Yu~Su.
\newblock Adaptive chameleon or stubborn sloth: Revealing the behavior of large language models in knowledge conflicts.
\newblock In \emph{The Twelfth International Conference on Learning Representations, {ICLR} 2024, Vienna, Austria, May 7-11, 2024}. OpenReview.net, 2024.
\newblock URL \url{https://openreview.net/forum?id=auKAUJZMO6}.

\bibitem[Xu et~al.(2024)Xu, Qi, Wang, Wang, Zhang, and Xu]{xu2024knowledge}
Rongwu Xu, Zehan Qi, Cunxiang Wang, Hongru Wang, Yue Zhang, and Wei Xu.
\newblock Knowledge conflicts for llms: A survey.
\newblock \emph{arXiv preprint arXiv:2403.08319}, 2024.

\bibitem[Zhao et~al.(2024{\natexlab{a}})Zhao, Khazanchi, Xing, He, Xu, and Lane]{zhao2024attacks}
Wanru Zhao, Vidit Khazanchi, Haodi Xing, Xuanli He, Qiongkai Xu, and Nicholas~Donald Lane.
\newblock Attacks on third-party apis of large language models.
\newblock \emph{arXiv preprint arXiv:2404.16891}, 2024{\natexlab{a}}.

\bibitem[Zhao et~al.(2024{\natexlab{b}})Zhao, Devoto, Hong, Du, Gema, Wang, He, Wong, and Minervini]{zhao2024steering}
Yu~Zhao, Alessio Devoto, Giwon Hong, Xiaotang Du, Aryo~Pradipta Gema, Hongru Wang, Xuanli He, Kam-Fai Wong, and Pasquale Minervini.
\newblock Steering knowledge selection behaviours in llms via sae-based representation engineering.
\newblock \emph{arXiv preprint arXiv:2410.15999}, 2024{\natexlab{b}}.

\bibitem[Zhong et~al.(2023)Zhong, Huang, Wettig, and Chen]{zhong2023poisoning}
Zexuan Zhong, Ziqing Huang, Alexander Wettig, and Danqi Chen.
\newblock Poisoning retrieval corpora by injecting adversarial passages.
\newblock \emph{arXiv preprint arXiv:2310.19156}, 2023.

\bibitem[Zhu and Li(2023)]{zhu2023physics}
Zeyuan~Allen Zhu and Yuanzhi Li.
\newblock Physics of language models: Part 3.1, knowledge storage and extraction.
\newblock \emph{arXiv preprint arXiv:2309.14316}, 2023.

\bibitem[Zou et~al.(2024)Zou, Geng, Wang, and Jia]{zou2024poisonedrag}
Wei Zou, Runpeng Geng, Binghui Wang, and Jinyuan Jia.
\newblock Poisonedrag: Knowledge poisoning attacks to retrieval-augmented generation of large language models.
\newblock \emph{arXiv preprint arXiv:2402.07867}, 2024.

\end{thebibliography}
\bibliographystyle{plainnat}


\appendix
\newpage
\clearpage

\section{Probing Model Training Settings}
\label{sec:probing-training}
For all probing experiments, we train the probing model with an L$_{1}$ norm regularisation. The training objective is $\mathcal{L}=-\log P(y=y_i) + \lambda \Vert W \Vert_1$, where we set $\lambda$ to $3\times10^{-4}$ and $y_i$ is the label. We train 20 times with different random seeds for each probing task, and we report the average and deviation in our experiments.
We split the training and test datasets for the probing tasks, ensuring no overlapping questions between them.

\section{More Experimental Results on Knowledge Conflict Probing}
\label{sec:more-conflict-probing}
\begin{figure}[h]
    \centering
    \begin{subfigure}[b]{0.32\textwidth}
        \centering
        \includegraphics[width=\linewidth]{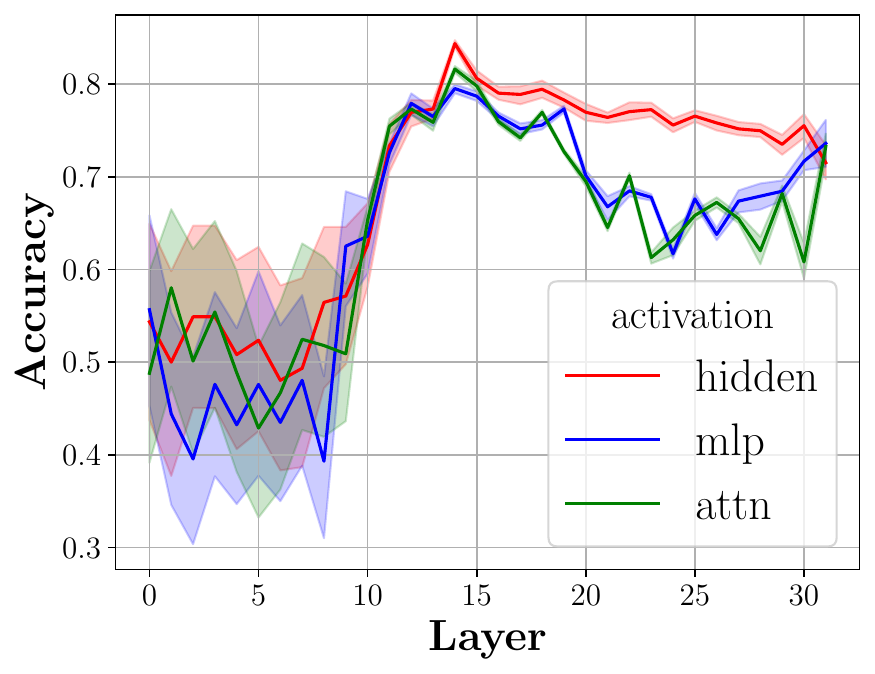}
    \end{subfigure}
    \begin{subfigure}[b]{0.32\textwidth}
        \centering
        \includegraphics[width=\linewidth]{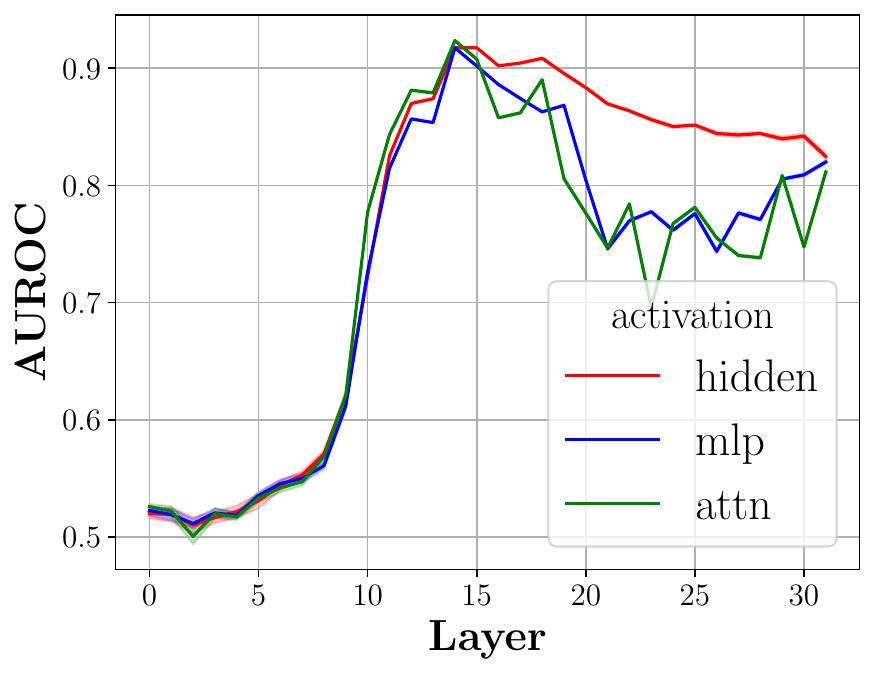}
    \end{subfigure}
    \begin{subfigure}[b]{0.32\textwidth}
        \centering
        \includegraphics[width=\linewidth]{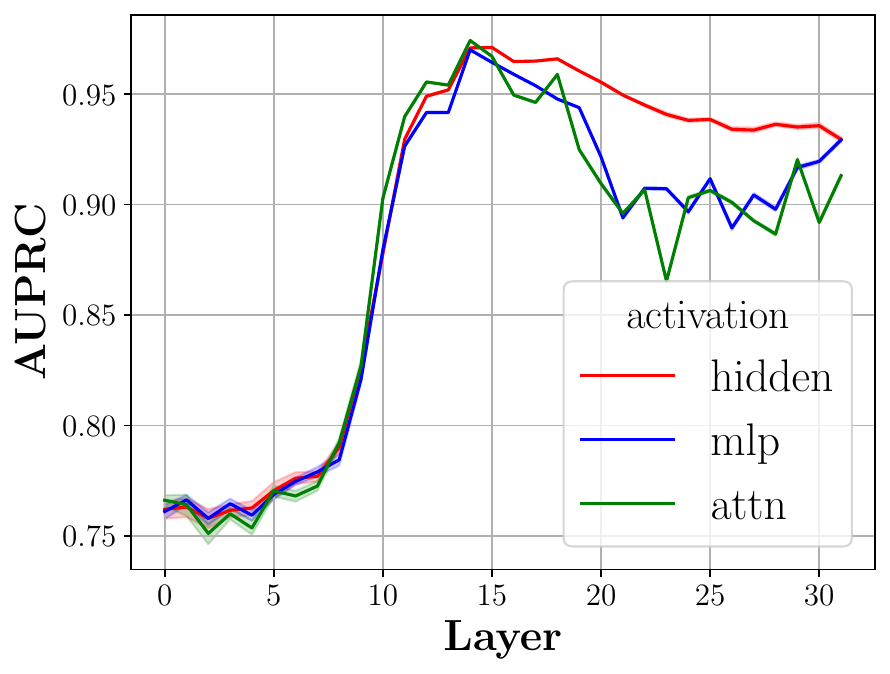}
    \end{subfigure}
\caption{Knowledge conflict probing results using Llama2-7B on NQSwap.}
\label{fig:knowledge-conflict-probing-llama2}
\end{figure}

\begin{figure}[h]
    \centering
    \begin{subfigure}[b]{0.32\textwidth}
        \centering
        \includegraphics[width=\linewidth]{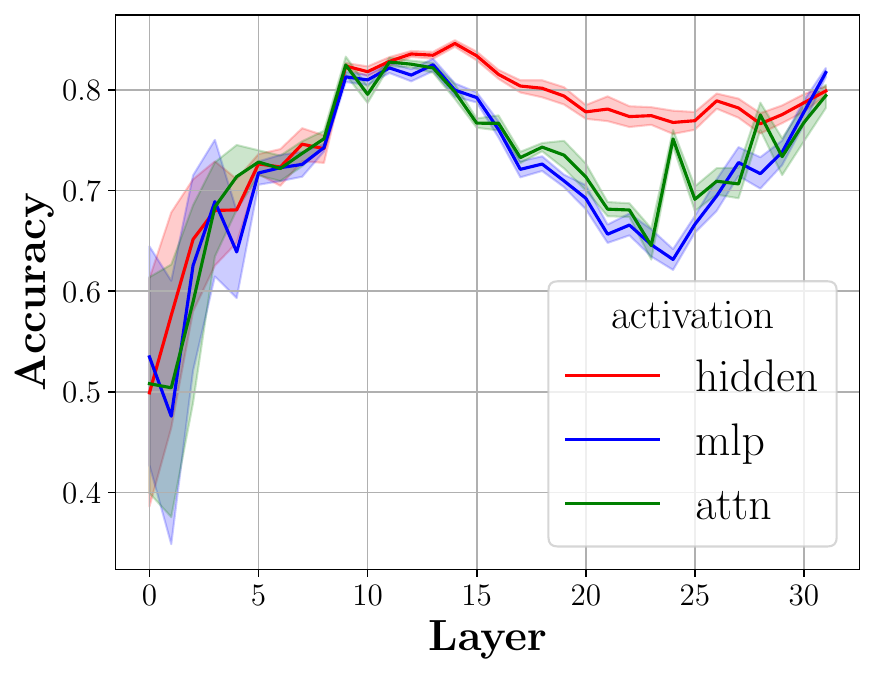}
    \end{subfigure}
    \begin{subfigure}[b]{0.32\textwidth}
        \centering
        \includegraphics[width=\linewidth]{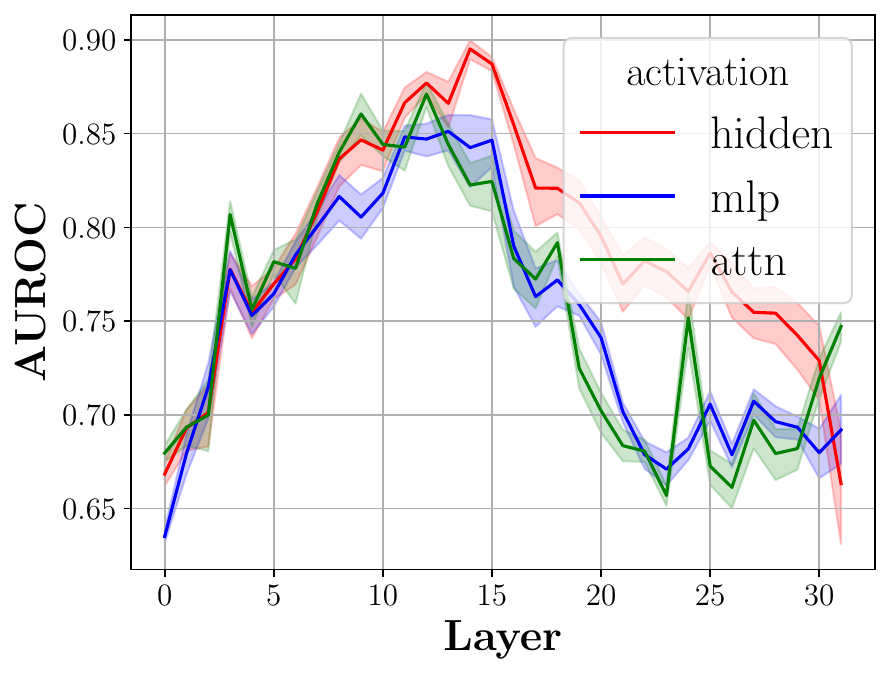}
    \end{subfigure}
    \begin{subfigure}[b]{0.32\textwidth}
        \centering
        \includegraphics[width=\linewidth]{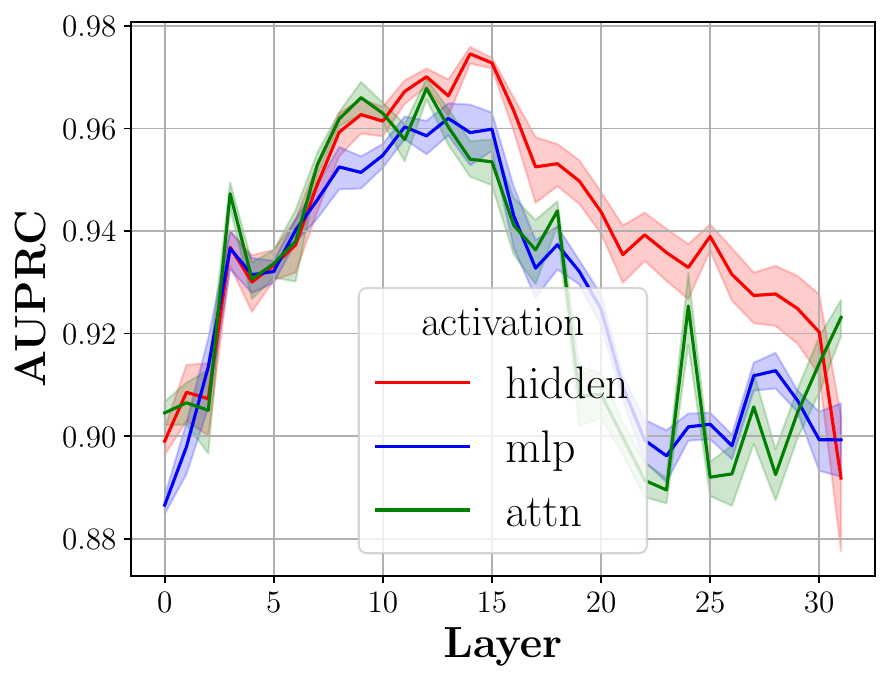}
    \end{subfigure}
\caption{Knowledge conflict probing results using Llama2-7B on Macnoise.}
\label{fig:knowledge-conflict-probing-llama2-macnoise}
\end{figure}

\begin{figure}[h]
    \centering
    \begin{subfigure}[b]{0.32\textwidth}
        \centering
        \includegraphics[width=\linewidth]{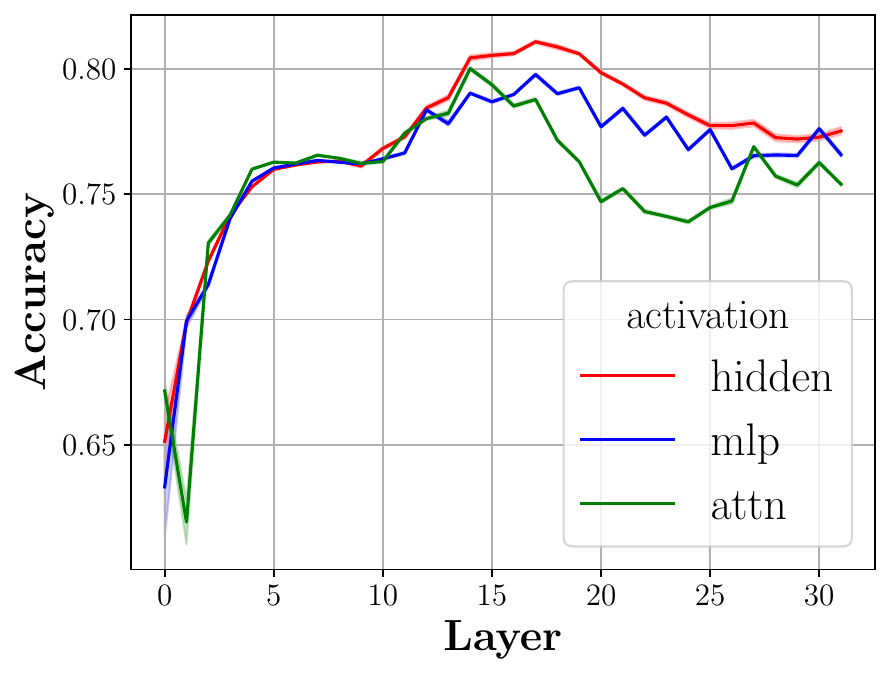}
    \end{subfigure}
    \begin{subfigure}[b]{0.32\textwidth}
        \centering
        \includegraphics[width=\linewidth]{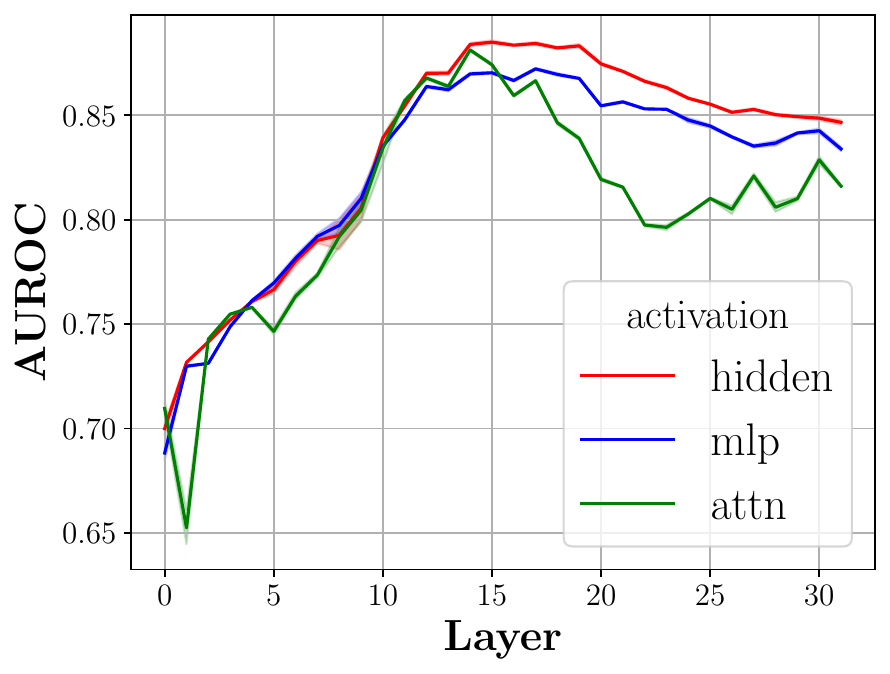}
    \end{subfigure}
    \begin{subfigure}[b]{0.32\textwidth}
        \centering
        \includegraphics[width=\linewidth]{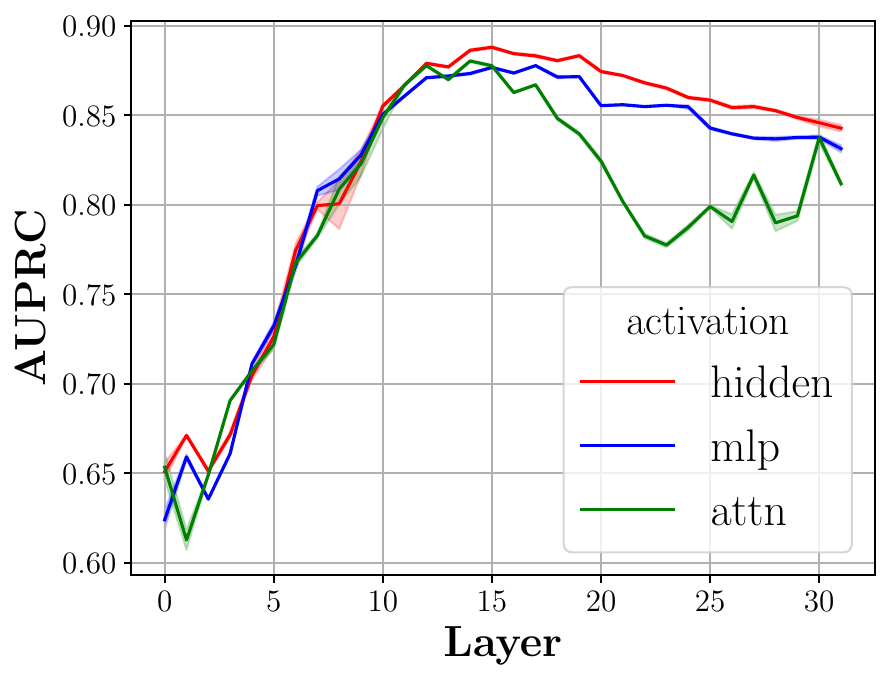}
    \end{subfigure}
\caption{Knowledge conflict probing results using Llama2-7B on ConflictQA.}
\label{fig:knowledge-conflict-probing-llama2-conflictqa}
\end{figure}

We present the knowledge conflict probing results on Macnoise, NQSwap, ConflictQA using Llama2-7B in~\cref{fig:knowledge-conflict-probing-llama2},~\cref{fig:knowledge-conflict-probing-llama2-macnoise} and~\cref{fig:knowledge-conflict-probing-llama2-conflictqa}.
The results match the trend discussed in~\cref{sec:results-and-findings}, where the model exhibits an internal mechanism for identifying conflicts. The signal of knowledge conflict peaks around the 13th to 14th layers and gradually decreases in the later layers.

\newpage
\clearpage
\section{More Analysis of Skewness Patterns of Residual Streams}
\label{sec:more-skewness-plots}
\begin{figure}[h]
    \centering
    \begin{subfigure}[b]{0.32\textwidth}
        \centering
        \includegraphics[width=\linewidth]{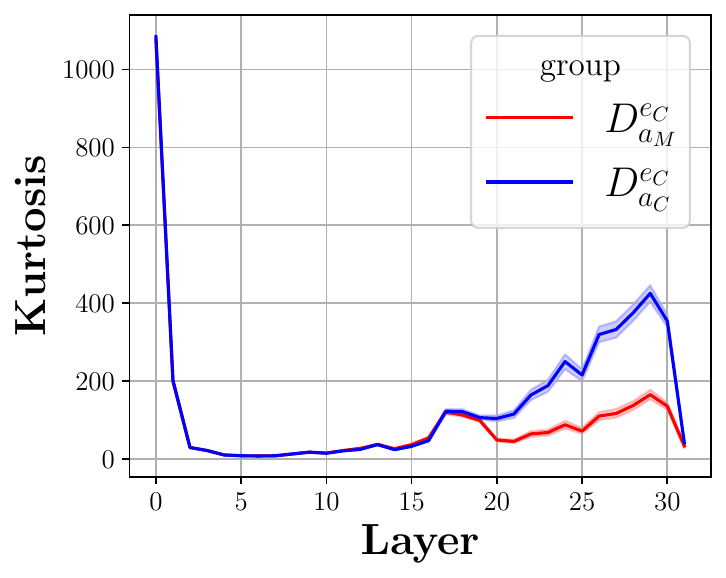}
    \end{subfigure}
    \begin{subfigure}[b]{0.32\textwidth}
        \centering
        \includegraphics[width=\linewidth]{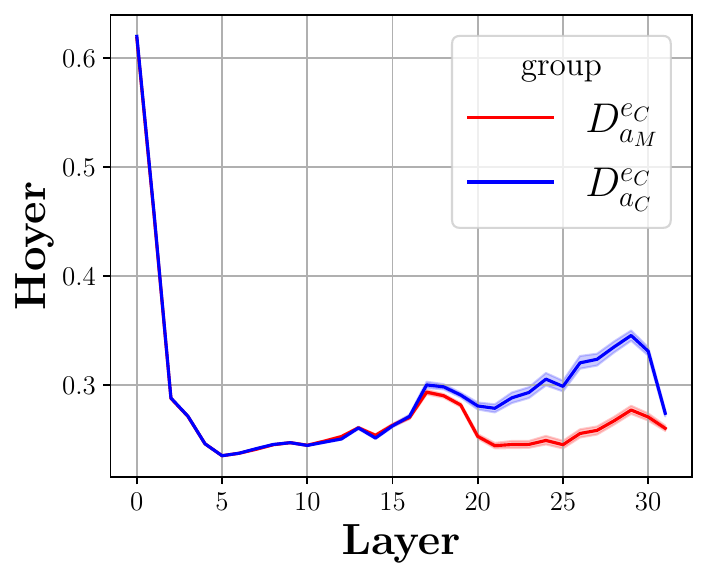}
    \end{subfigure}
    \begin{subfigure}[b]{0.32\textwidth}
        \centering
        \includegraphics[width=\linewidth]{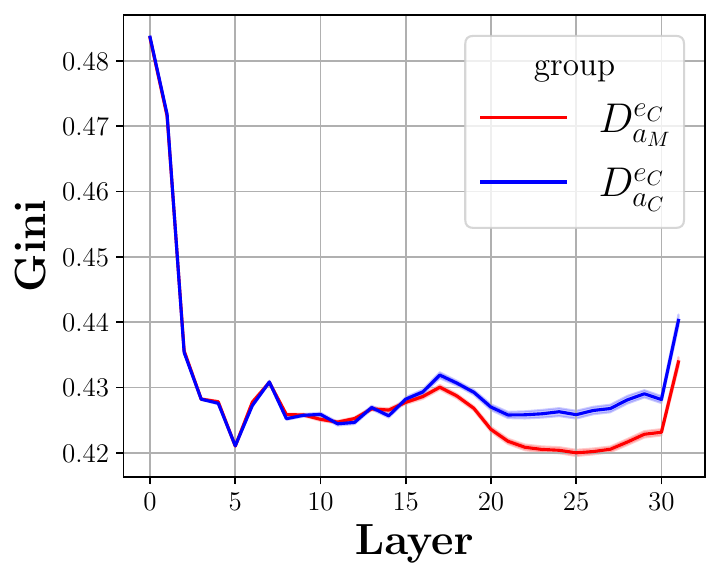}
    \end{subfigure}
\caption{Skewness of the hidden states of Llama2-7B on NQSwap.}
\label{fig:llama2-hidden-skewness}
\end{figure}

\begin{figure}[h]
    \centering
    \begin{subfigure}[b]{0.32\textwidth}
        \centering
        \includegraphics[width=\linewidth]{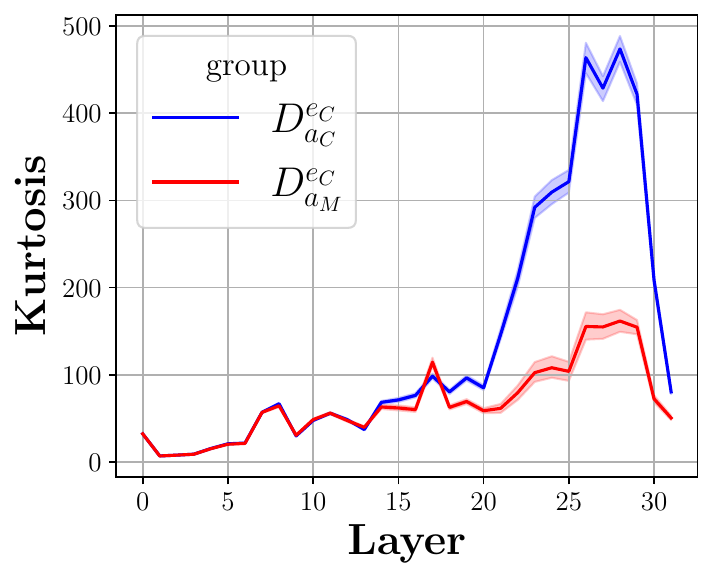}
    \end{subfigure}
    \begin{subfigure}[b]{0.32\textwidth}
        \centering
        \includegraphics[width=\linewidth]{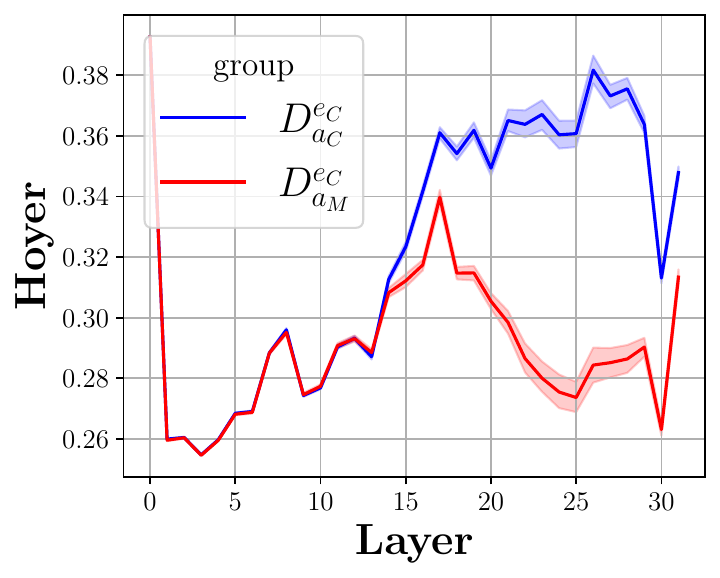}
    \end{subfigure}
    \begin{subfigure}[b]{0.32\textwidth}
        \centering
        \includegraphics[width=\linewidth]{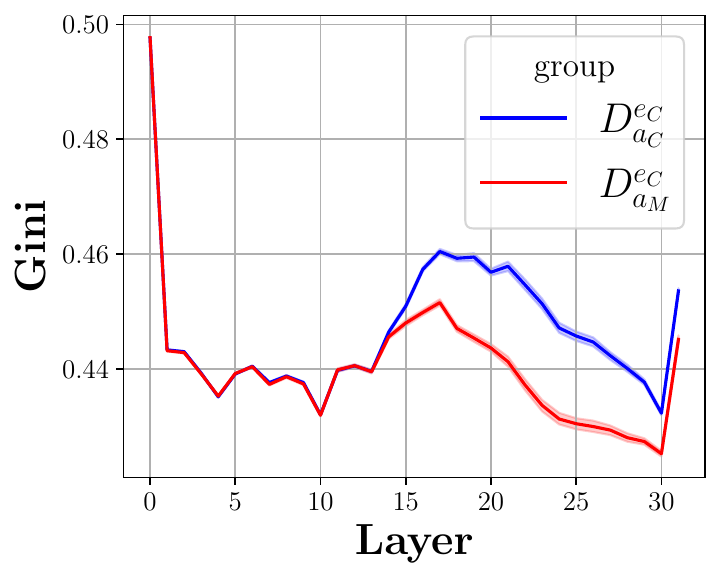}
    \end{subfigure}
\caption{Skewness of the hidden states of Llama3-8B on Macnoise.}
\label{fig:llama3-hidden-skewness-macnoise}
\end{figure}

\begin{figure}[h]
    \centering
    \begin{subfigure}[b]{0.32\textwidth}
        \centering
        \includegraphics[width=\linewidth]{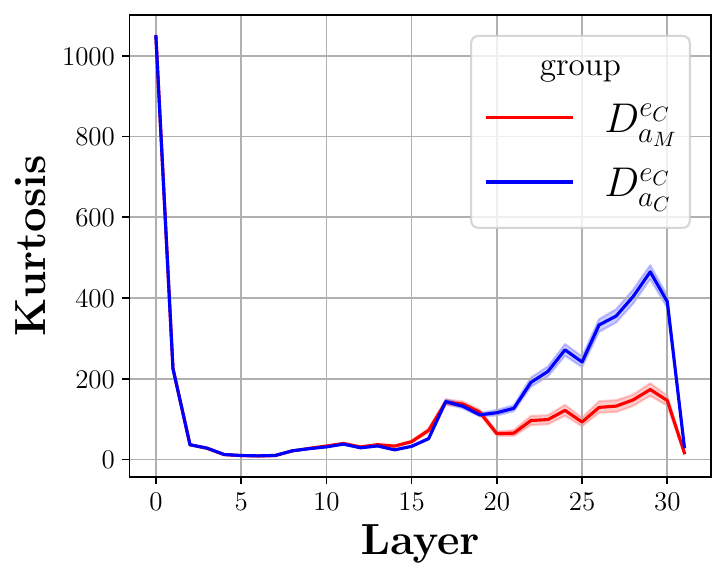}
    \end{subfigure}
    \begin{subfigure}[b]{0.32\textwidth}
        \centering
        \includegraphics[width=\linewidth]{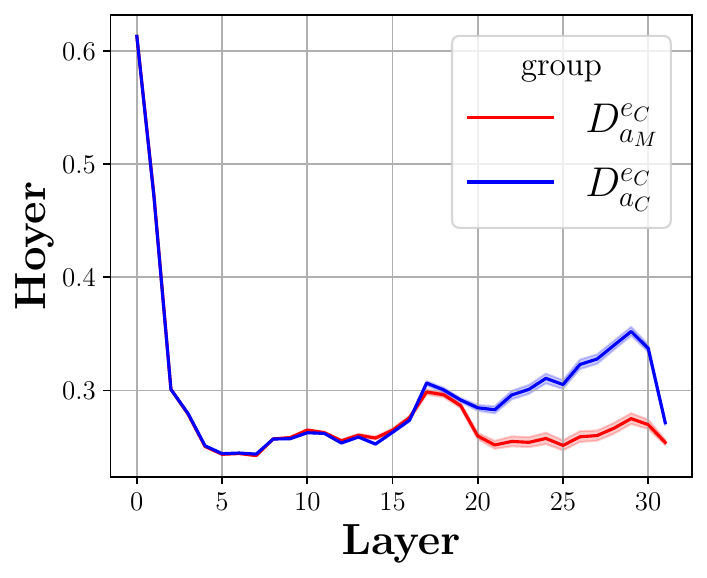}
    \end{subfigure}
    \begin{subfigure}[b]{0.32\textwidth}
        \centering
        \includegraphics[width=\linewidth]{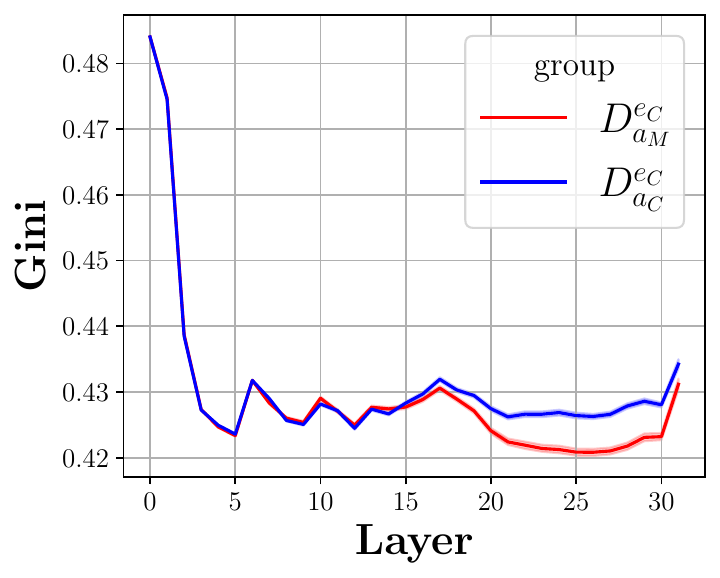}
    \end{subfigure}
\caption{Skewness of the hidden states of Llama2-7B on Macnoise.}
\label{fig:llama2-hidden-skewness-macnoise}
\end{figure}

\begin{figure}[t]
    \centering
    \begin{subfigure}[b]{0.32\textwidth}
        \centering
        \includegraphics[width=\linewidth]{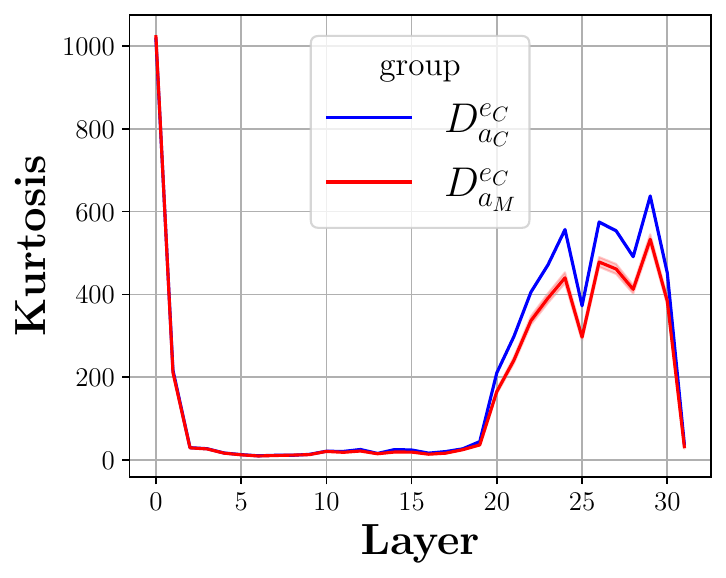}
    \end{subfigure}
    \begin{subfigure}[b]{0.32\textwidth}
        \centering
        \includegraphics[width=\linewidth]{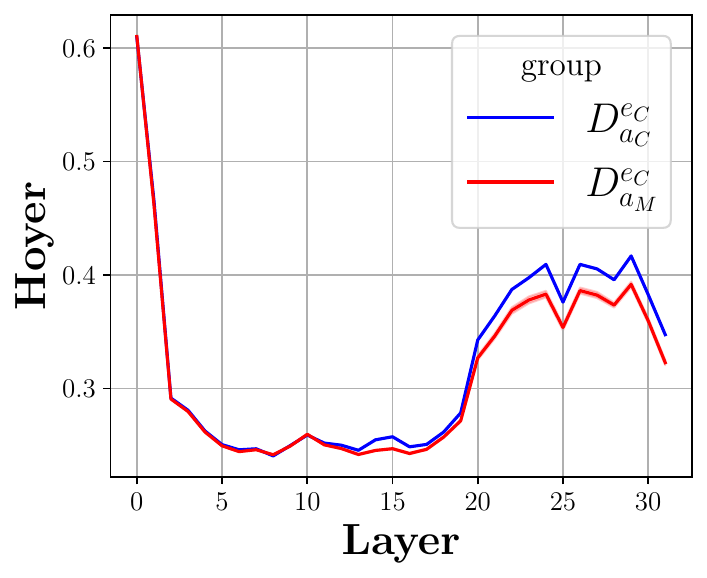}
    \end{subfigure}
    \begin{subfigure}[b]{0.32\textwidth}
        \centering
        \includegraphics[width=\linewidth]{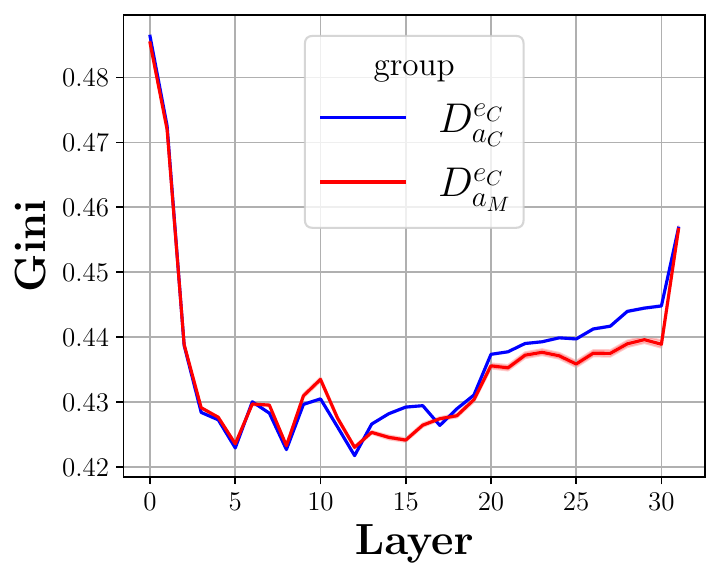}
    \end{subfigure}
\caption{
Skewness of the hidden states of Llama-27B on ConflictQA.
}
\label{fig:kurtosis-llama2-conflictqa}
\end{figure}

We present the skewness of the hidden state of Llama2-7B on NQSwap in~\cref{fig:llama2-hidden-skewness}.
It shows the same pattern as we discussed in~\cref{fig:kurtosis-llama3}, where the residual stream exhibits significantly more skewed distribution when using contextual knowledge compared with using parametric knowledge from the 17th layer.
In addition to NQSwap, we analyse the skewness pattern using Macnoise~\citep{macnoise} and ConflictQA~\citep{conflictqa}.
As shown in~\cref{fig:llama3-hidden-skewness-macnoise}, \cref{fig:llama2-hidden-skewness-macnoise}, \cref{fig:kurtosis-llama2-conflictqa}, we find that the model also shows a similar skewness pattern with NQSwap, where the residual stream exhibits a more skewed distribution from middle layers when the model uses the contextual knowledge.

\begin{figure}[t]
    \centering
    \begin{subfigure}[b]{0.32\textwidth}
        \centering
        \includegraphics[width=\linewidth]{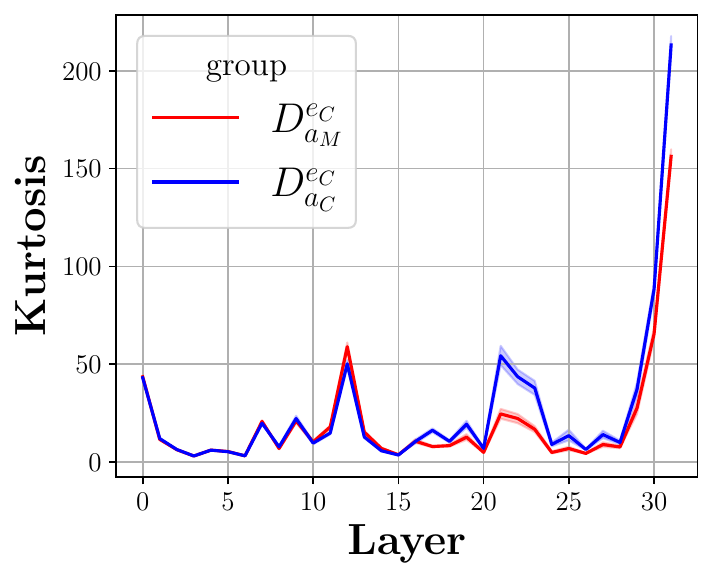}
    \end{subfigure}
    \begin{subfigure}[b]{0.32\textwidth}
        \centering
        \includegraphics[width=\linewidth]{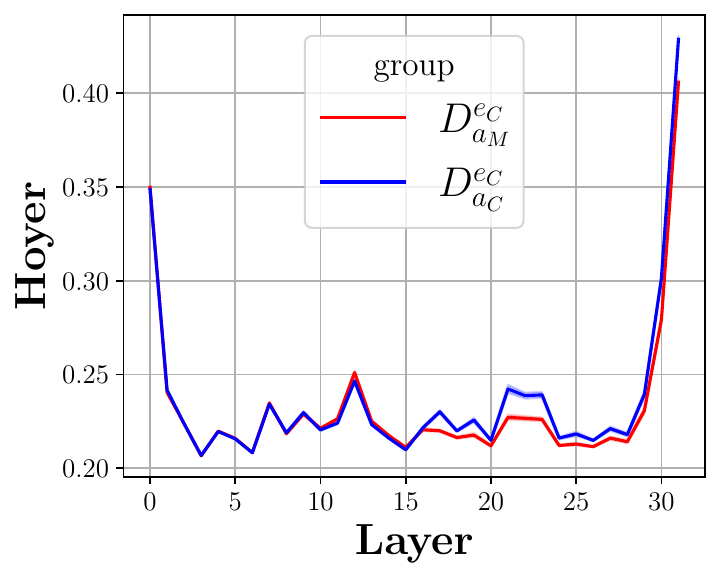}
    \end{subfigure}
    \begin{subfigure}[b]{0.32\textwidth}
        \centering
        \includegraphics[width=\linewidth]{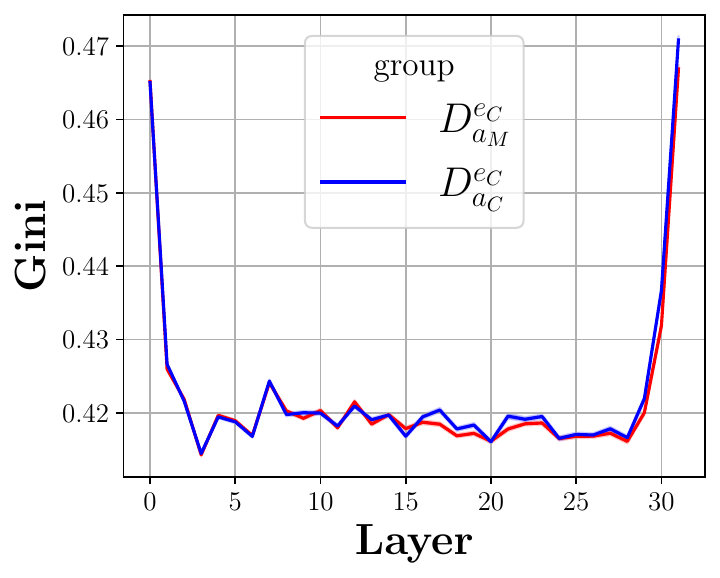}
    \end{subfigure}
\caption{Skewness of the MLP activation of Llama3-8B on NQSwap.}
\label{fig:llama3-mlp-skewness}
\end{figure}

\begin{figure}[t]
    \centering
    \begin{subfigure}[b]{0.32\textwidth}
        \centering
        \includegraphics[width=\linewidth]{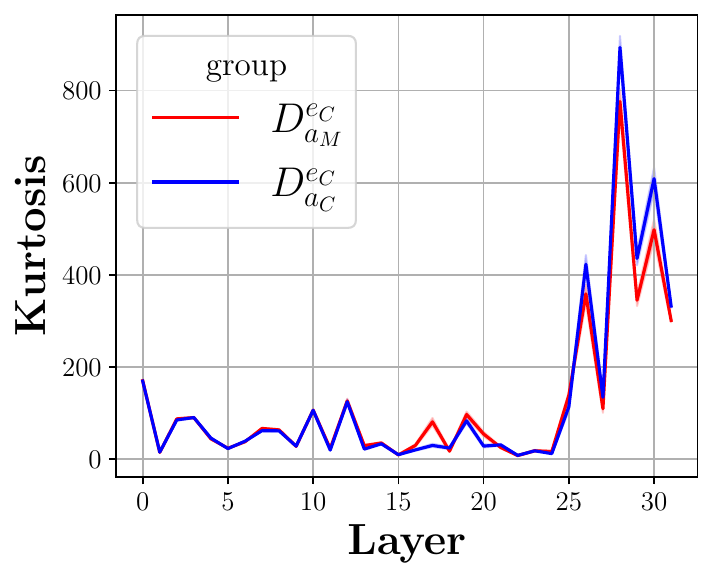}
    \end{subfigure}
    \begin{subfigure}[b]{0.32\textwidth}
        \centering
        \includegraphics[width=\linewidth]{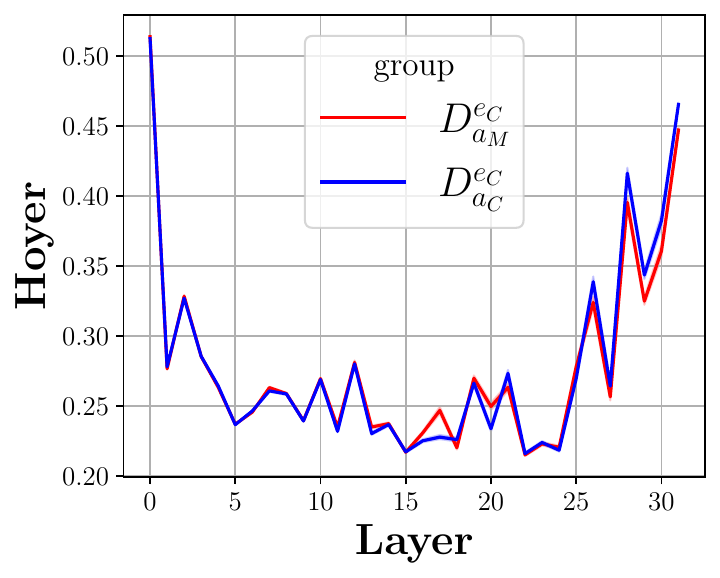}
    \end{subfigure}
    \begin{subfigure}[b]{0.32\textwidth}
        \centering
        \includegraphics[width=\linewidth]{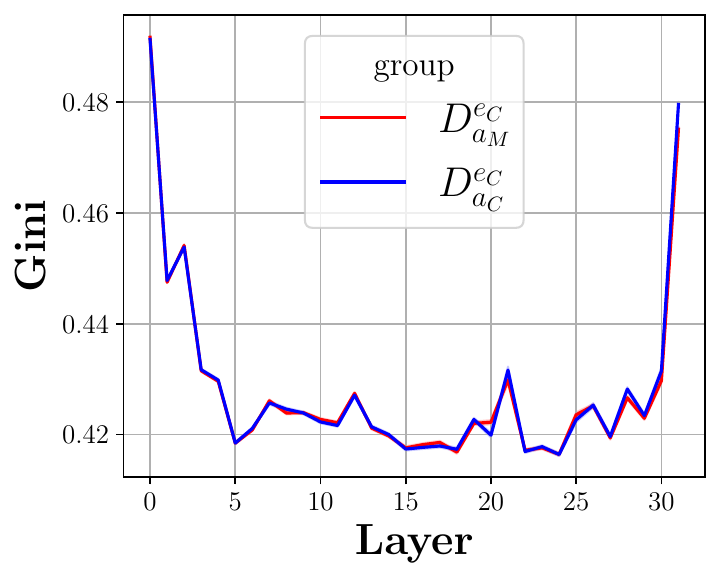}
    \end{subfigure}
\caption{Skewness of the Self-Attention activation of Llama3-8B on NQSwap.}
\label{fig:llama3-attn-skewness}
\end{figure}

\begin{figure}[t]
    \centering
    \begin{subfigure}[b]{0.32\textwidth}
        \centering
        \includegraphics[width=\linewidth]{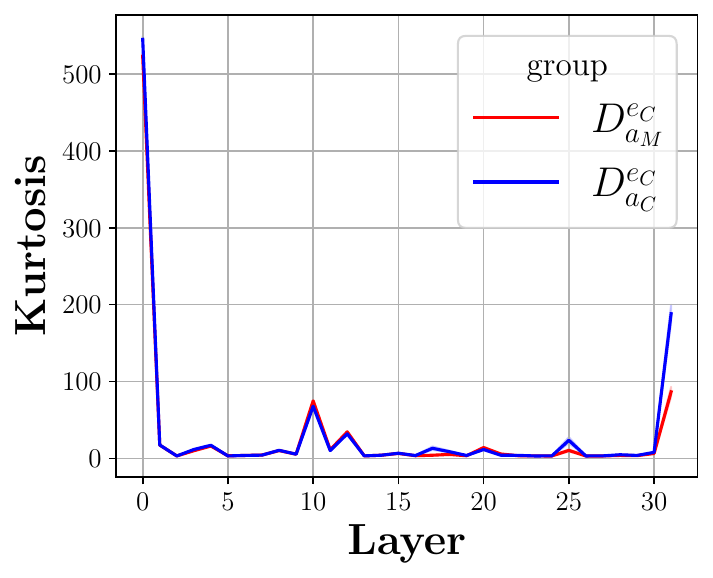}
    \end{subfigure}
    \begin{subfigure}[b]{0.32\textwidth}
        \centering
        \includegraphics[width=\linewidth]{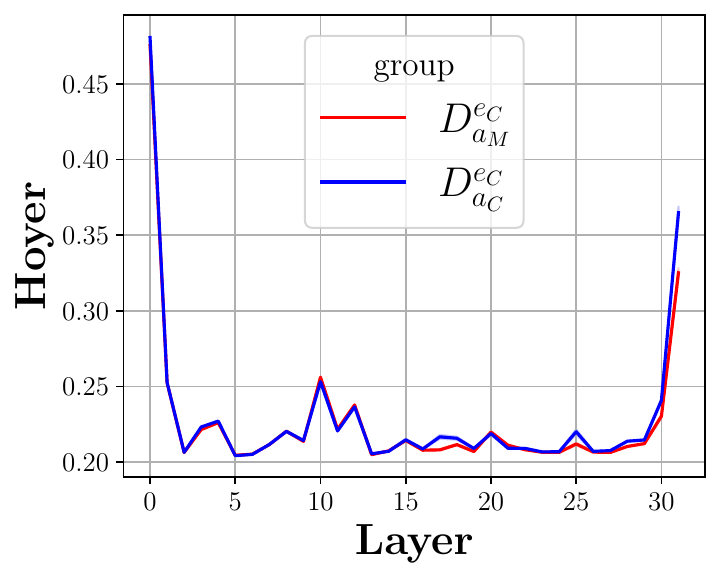}
    \end{subfigure}
    \begin{subfigure}[b]{0.32\textwidth}
        \centering
        \includegraphics[width=\linewidth]{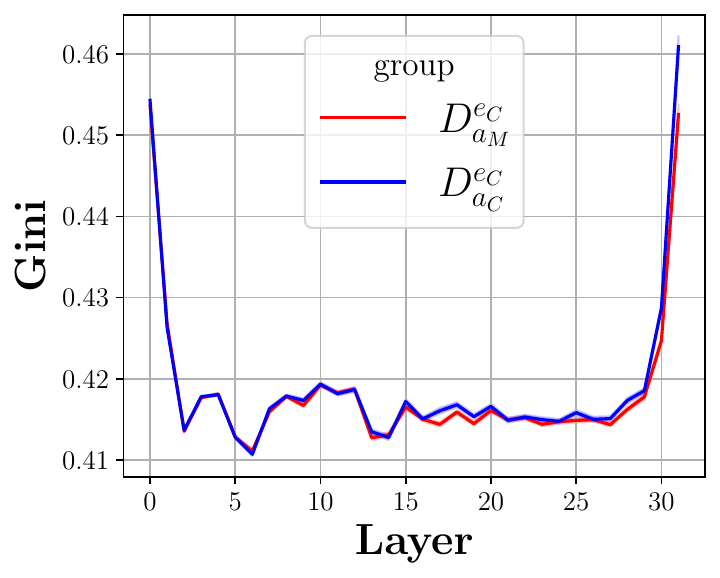}
    \end{subfigure}
\caption{Skewness of the MLP activation of Llama2-7B on NQSwap.}
\label{fig:kurtosis-llama2-mlp-skewness}
\end{figure}

\begin{figure}[t]
    \centering
    \begin{subfigure}[b]{0.32\textwidth}
        \centering
        \includegraphics[width=\linewidth]{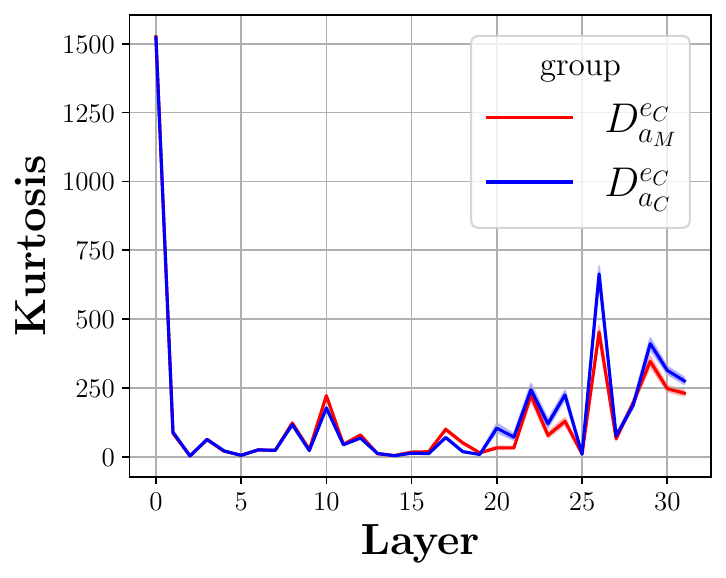}
    \end{subfigure}
    \begin{subfigure}[b]{0.32\textwidth}
        \centering
        \includegraphics[width=\linewidth]{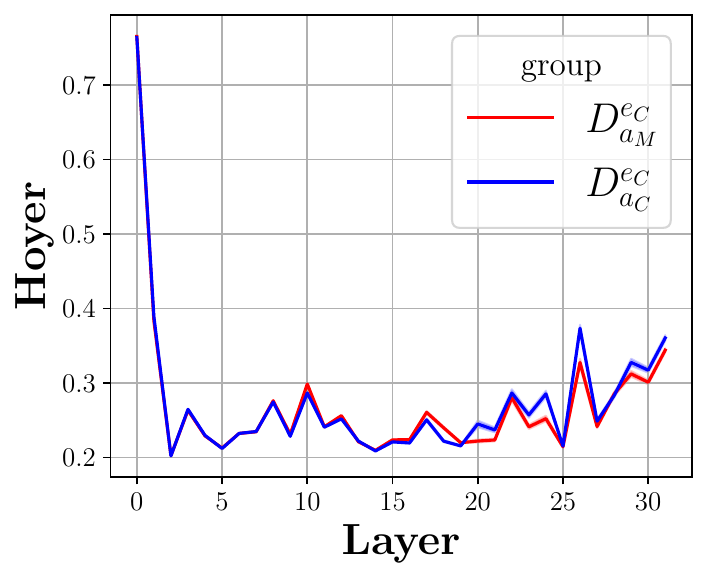}
    \end{subfigure}
    \begin{subfigure}[b]{0.32\textwidth}
        \centering
        \includegraphics[width=\linewidth]{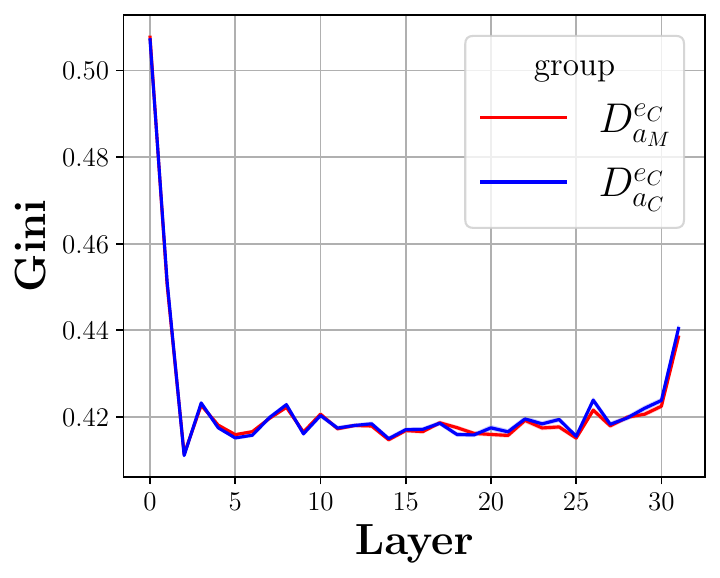}
    \end{subfigure}
\caption{Skewness of the Self-Attention activation of Llama2-7B on NQSwap.}
\label{fig:kurtosis-llama2-attn-skewness}
\end{figure}

We also analyse the skewness of MLP and Self-Attention activations, presented in~\cref{fig:llama3-mlp-skewness},~\cref{fig:llama3-attn-skewness},~\cref{fig:kurtosis-llama2-mlp-skewness}, and~\cref{fig:kurtosis-llama2-attn-skewness}.
However, we do not observe a specific skewness pattern in MLP and Self-Attention activations.

\newpage
\clearpage
\section{L1 Norm and L2 Norm Values of Residual Streams}
\label{sec:more-distribution-patterns}
\begin{figure}[h]
    \centering
    \begin{subfigure}[b]{0.32\textwidth}
        \centering
        \includegraphics[width=\linewidth]{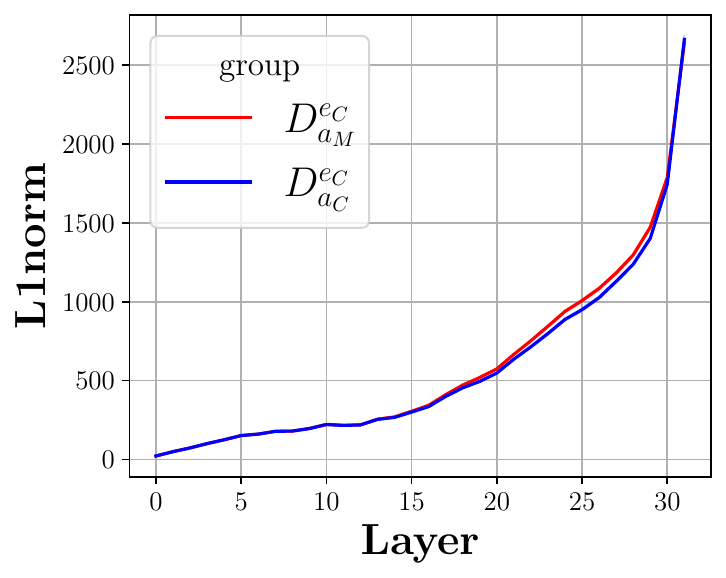}
    \end{subfigure}
    \begin{subfigure}[b]{0.32\textwidth}
        \centering
        \includegraphics[width=\linewidth]{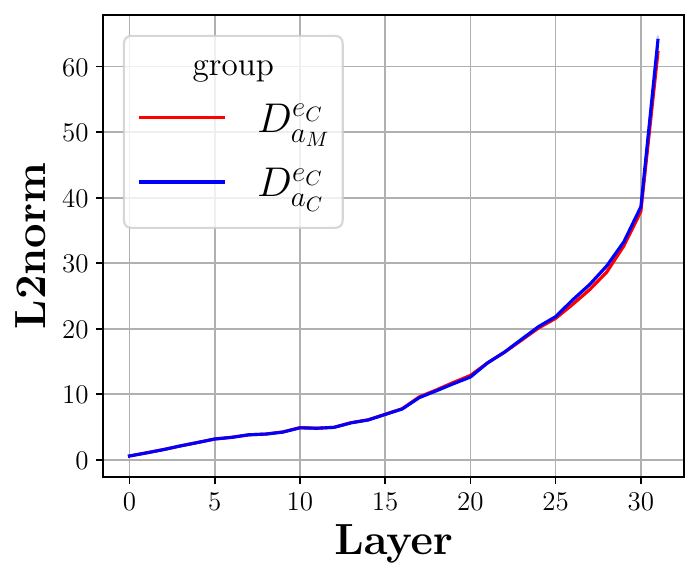}
    \end{subfigure}
\caption{L1 norm and L2 norm of the hidden states of Llama3-8B on NQSwap.}
\label{fig:llama3-norm}
\end{figure}

\begin{figure}[h]
    \centering
    \begin{subfigure}[b]{0.32\textwidth}
        \centering
        \includegraphics[width=\linewidth]{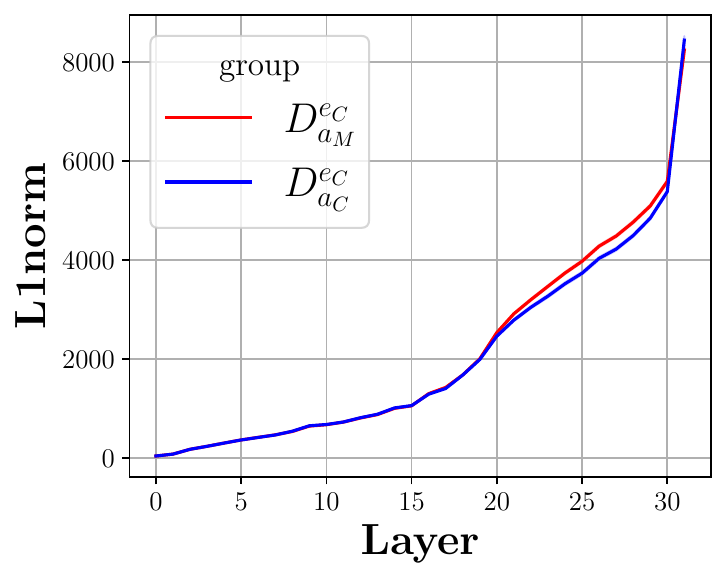}
    \end{subfigure}
    \begin{subfigure}[b]{0.32\textwidth}
        \centering
        \includegraphics[width=\linewidth]{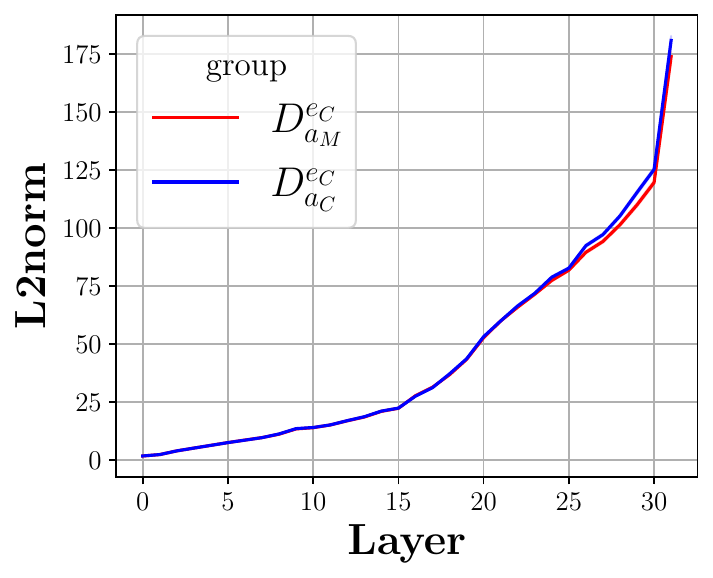}
    \end{subfigure}
\caption{L1 norm and L2 norm of the hidden states of Llama2-7B on NQSwap.}
\label{fig:llama2-norm}
\end{figure}

We present L1 Norm and L2 Norm of the residual stream in the~\cref{fig:llama3-norm} and~\cref{fig:llama2-norm}.
We found that though the residual stream show distinct skewness patterns in $D_{a_C}^{e_C}$ and $D_{a_M}^{e_C}$, the L1 norm and L2 norm of the them do not have a significant difference.

\newpage
\clearpage
\section{More Experimental Results on Knowledge Selection Probing}
\label{sec:more-behaviour}

\begin{figure}[h]
    \centering
    \begin{subfigure}[b]{0.32\textwidth}
        \centering
        \includegraphics[width=\linewidth]{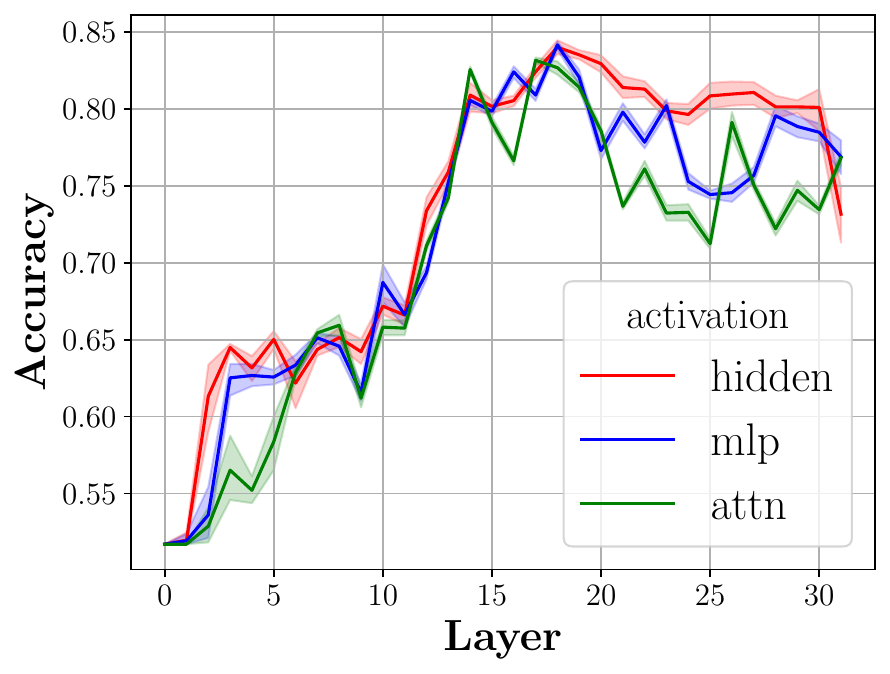}
    \end{subfigure}
    \begin{subfigure}[b]{0.32\textwidth}
        \centering
        \includegraphics[width=\linewidth]{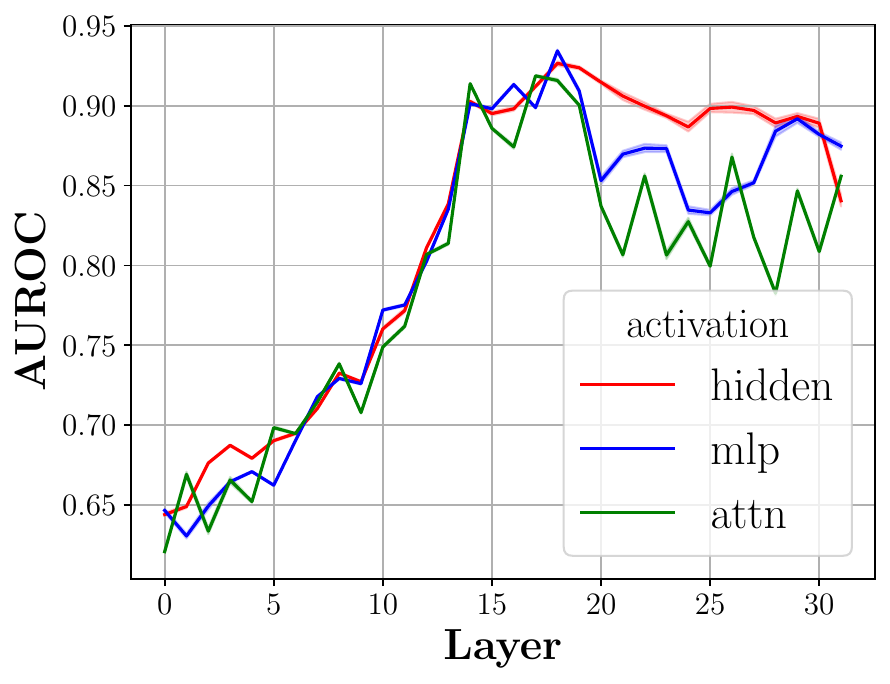}
    \end{subfigure}
    \begin{subfigure}[b]{0.32\textwidth}
        \centering
        \includegraphics[width=\linewidth]{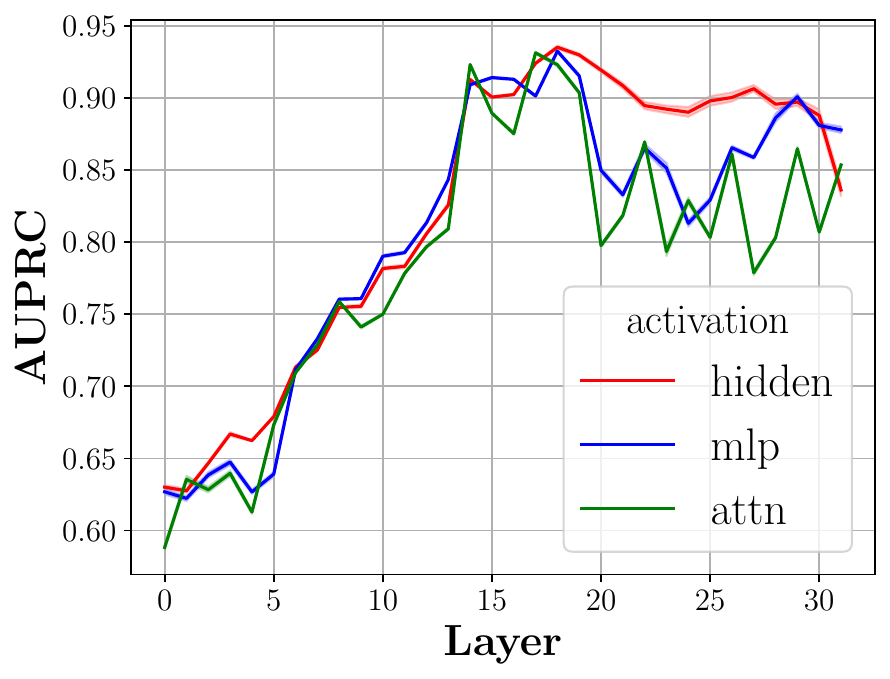}
    \end{subfigure}
\caption{Knowledge selection probing results using Llama2-7B on NQSwap.
}
\label{fig:behaviour-probing-llama2}
\end{figure}

\begin{figure}[h]
    \centering
    \begin{subfigure}[b]{0.32\textwidth}
        \centering
        \includegraphics[width=\linewidth]{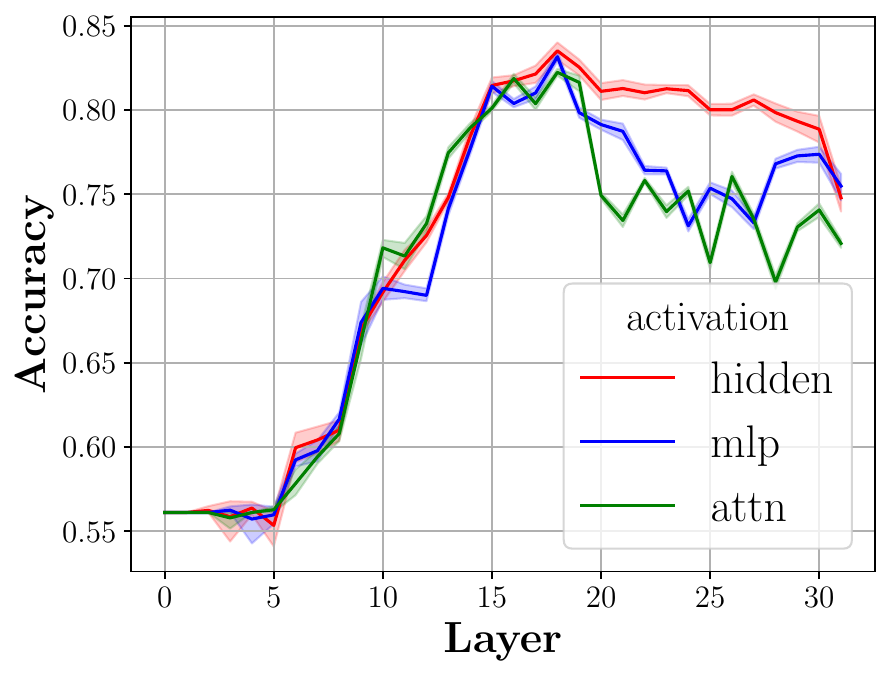}
    \end{subfigure}
    \begin{subfigure}[b]{0.32\textwidth}
        \centering
        \includegraphics[width=\linewidth]{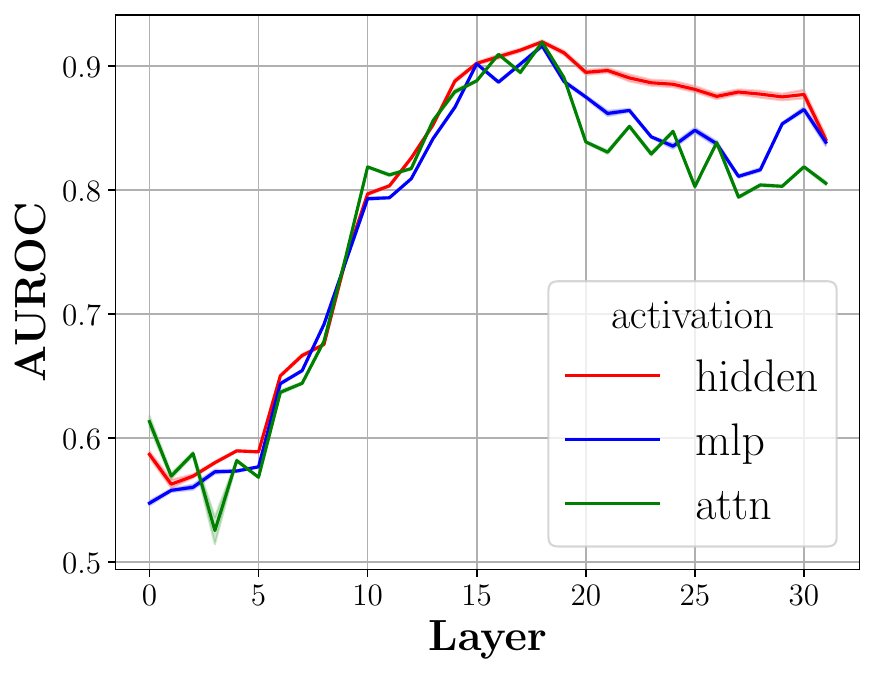}
    \end{subfigure}
    \begin{subfigure}[b]{0.32\textwidth}
        \centering
        \includegraphics[width=\linewidth]{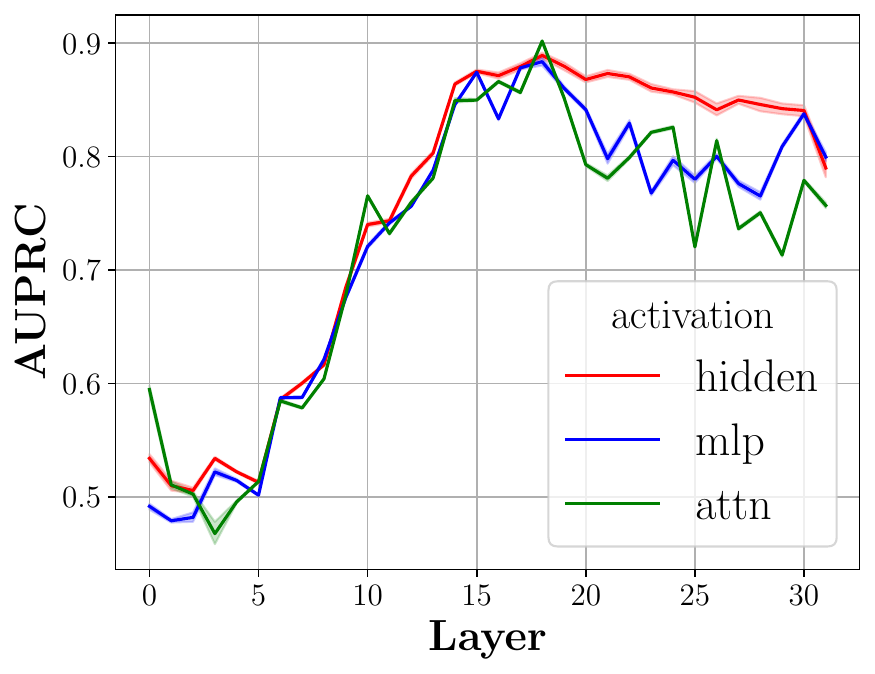}
    \end{subfigure}
\caption{Knowledge selection probing results using Llama2-7B on Macnoise.
}
\label{fig:behaviour-probing-llama2-macnoise}
\end{figure}

\begin{figure}[h]
    \centering
    \begin{subfigure}[b]{0.32\textwidth}
        \centering
        \includegraphics[width=\linewidth]{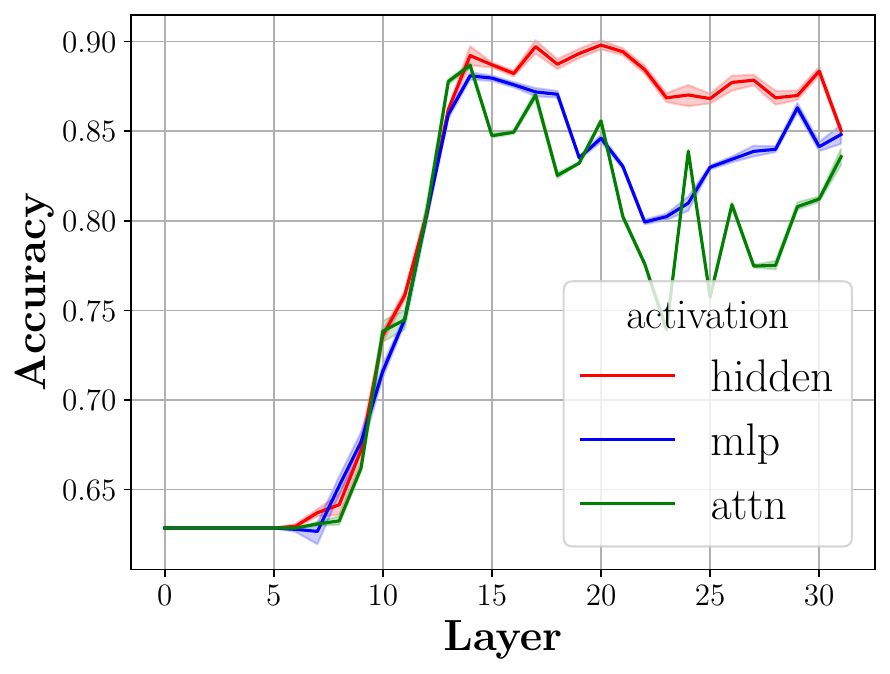}
    \end{subfigure}
    \begin{subfigure}[b]{0.32\textwidth}
        \centering
        \includegraphics[width=\linewidth]{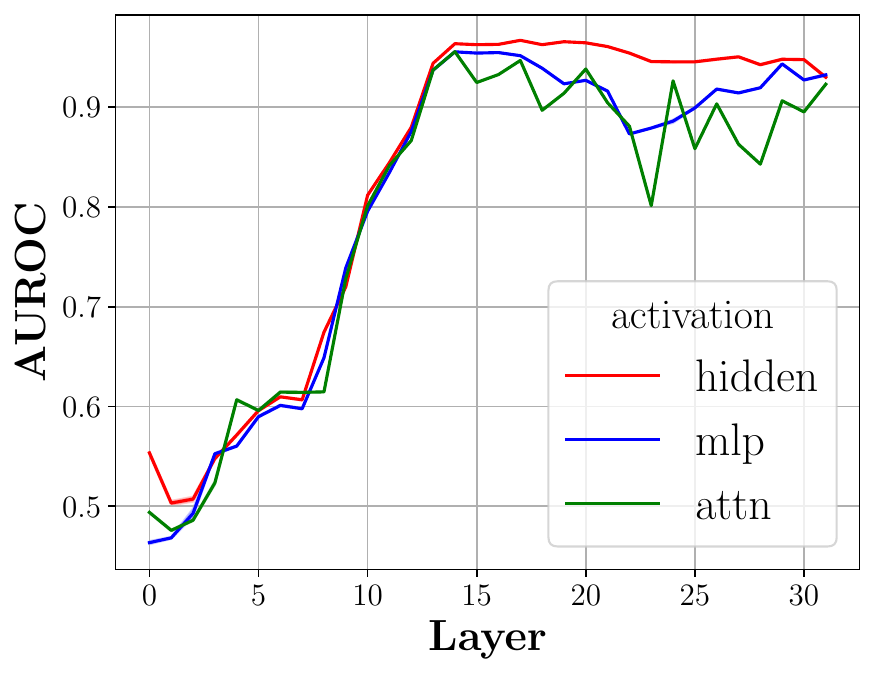}
    \end{subfigure}
    \begin{subfigure}[b]{0.32\textwidth}
        \centering
        \includegraphics[width=\linewidth]{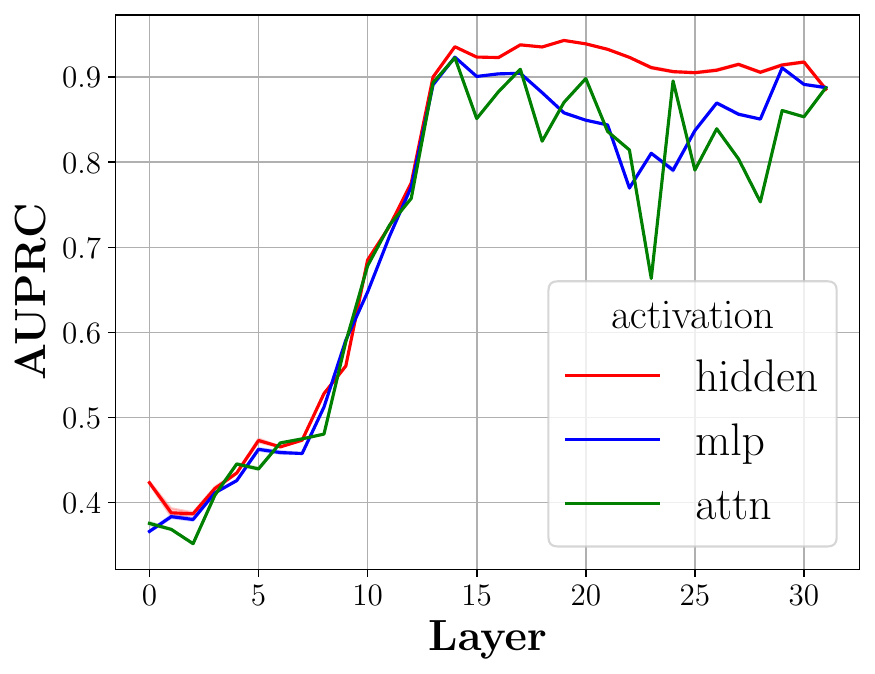}
    \end{subfigure}
\caption{Knowledge selection probing results using Llama3-8B on Macnoise.
}
\label{fig:behaviour-probing-llama3-macnoise}
\end{figure}

We present additional knowledge selection probing results on NQSwap and Macnoise using Llama2-7B and Llama3-8B in~\cref{fig:behaviour-probing-llama2},~\cref{fig:behaviour-probing-llama2-macnoise} and~\cref{fig:behaviour-probing-llama3-macnoise}.
The results show a similar trend as shown in~\cref{fig:behaviour-probing-llama3}, where the probing model reaches the highest accuracy at around the 17th layer, which is later than the aggregation of knowledge conflict signal at the 14th layer.

\end{document}